\documentclass{article}

\PassOptionsToPackage{numbers, compress}{natbib}

\usepackage[final]{neurips_2025}

\usepackage[utf8]{inputenc} 
\usepackage[T1]{fontenc}    
\usepackage{hyperref}       
\usepackage{url}            
\usepackage{booktabs}       
\usepackage{amsfonts}       
\usepackage{nicefrac}       
\usepackage{microtype}      
\usepackage{xcolor}         

\usepackage{subfiles}
\usepackage{abbreviations}
\usepackage{amsmath}
\usepackage{graphicx}
\usepackage{algorithm}
\usepackage{algpseudocode}
\usepackage[capitalize,noabbrev]{cleveref}
\usepackage{mathtools}
\usepackage{dsfont}
\usepackage{subfigure}

\title{Thompson Sampling for Multi-Objective Linear Contextual Bandit}

\author{
  Somangchan Park \\
  Seoul National University\\
  \texttt{jrhopefulp@snu.ac.kr} \\
  \And
  Heesang Ann \\
  Seoul National University \\
  \texttt{sang3798@snu.ac.kr} \\
  \And
  Min-hwan Oh \\
  Seoul National University \\
  \texttt{minoh@snu.ac.kr} \\
}

\begin{document}

\maketitle

\begin{abstract}
We study the multi-objective linear contextual bandit problem, where multiple possible conflicting objectives must be optimized simultaneously. We propose \texttt{MOL-TS}, the \textit{first} Thompson Sampling algorithm with Pareto regret guarantees for this problem. Unlike standard approaches that compute an empirical Pareto front each round, \texttt{MOL-TS} samples parameters across objectives and efficiently selects an arm from a novel \emph{effective Pareto front}, which accounts for repeated selections over time. Our analysis shows that \texttt{MOL-TS} achieves a worst-case Pareto regret bound of $\widetilde{O}(d^{3/2}\sqrt{T})$, where $d$ is the dimension of the feature vectors, $T$ is the total number of rounds, matching the best known order for randomized linear bandit algorithms for single objective. Empirical results confirm the benefits of our proposed approach, demonstrating improved regret minimization and strong multi-objective performance.
\end{abstract}
\section{Introduction}\label{sec:introduction}

The multi-objective multi-armed bandit (MOMAB) problem~\citep{drugan2013momab, auer2016pareto,tekin2018multi,turgay2018multi, lu2019moglb, xu2023pareto, Kim2023, cheng2024hierarchize} generalizes the classical, single-objective multi-armed bandit to settings with multiple, potentially conflicting objectives. Pulling an arm yields a \emph{vector} of objective-specific rewards, so a single “best” arm is often ill-defined and optimality must account for trade-offs across objectives.

One simple way to handle trade-offs is \emph{scalarization}~\citep{paria2020flexible, zhang2020random, zhang2024}, which maps reward vectors to a scalar via, e.g., weighted sums, minimax, or other nonlinear transforms, thereby reducing MOMAB to a single-objective bandit. However, selecting a suitable scalarization is itself nontrivial, and an arm optimal under one scalarization can be markedly suboptimal under another. An alternative is to reason directly in the vector space via \emph{Pareto optimality}~\citep{drugan2013momab, auer2016pareto, lu2019moglb, xu2023pareto, Kim2023, cheng2024hierarchize}: the Pareto front comprises arms whose mean reward vectors are not dominated component-wise. Because it compares reward vectors objective-wise, Pareto optimality is strictly more general than committing to a fixed scalarization, and we adopt it throughout.

All of the prior MOMAB work in Pareto regret propose and study UCB-based algorithms~\citep{drugan2013momab, lu2019moglb, xu2023pareto, Kim2023}. To the best of our knowledge, the Pareto regret of Thompson Sampling (TS)~\citep{agrawal2013thompson, abeille2017linear} has not been studied for MOMAB. 
This gap is noteworthy: in many single-objective contextual and non-contextual bandits, TS and its variants are empirically competitive or superior to UCB methods~\citep{thompson1933likelihood, chapelle2011empirical, agrawal2013thompson, abeille2017linear}, yet their worst-case analyses are typically more delicate. 
Extending TS to MOMAB introduces additional challenges, including coordinating randomized samples across multiple objectives and handling the possibility of multiple Pareto-optimal arms.

In this work, we develop a TS algorithm for the multi-objective linear contextual bandit and analyze its worst-case Pareto regret. Our method samples separate parameters for each objective and evaluates arms via an \emph{optimistic sampling} mechanism that ensures a nontrivial probability of being jointly optimistic across all objectives (Section~\ref{sec:opt-sampling}), which underpins the regret analysis.

We also revisit the notion of performance measurement. The standard Pareto regret compares per-round mean reward vectors; consequently, repeatedly playing a single Pareto-optimal arm can yield zero regret even when an alternative arm selection would strictly larger \emph{cumulative} rewards (see \cref{section:p}). To address this, we introduce the notion of an \emph{effective Pareto-optimal arm} (\cref{def:CPO}): an arm that is Pareto-optimal per round and remains undominated when evaluated through the lens of cumulative rewards under any number of repeated selections. Building on this, we define a corresponding regret notion that penalizes policies vulnerable to such cumulative inefficiencies; our algorithm targets the \emph{effective} Pareto front to avoid these failures.

Our main contributions are summarized as follows:
\begin{itemize}
    \item We propose an algorithm for the multi-objective linear contextual bandit problem: \textit{Thompson sampling for Multi-objective Linear Bandit} (\texttt{MOL-TS}).
    To the best of our knowledge, 
    this is the first randomized algorithm for multi-objective contextual bandits with Pareto regret guarantees.
    Unlike the existing multi-objective algorithms, \texttt{MOL-TS} does not explicitly compute an empirical Pareto front each round, but rather randomly selects an arm from that Pareto front, which is much more computationally efficiently. 

    \item We propose the concept of a \textit{effective Pareto optimal arm}~(\cref{def:CPO}), which satisfies the condition of Pareto optimal arm, and also the total rewards for every objective with any number of its repeated selection satisfying Pareto optimality. Any arm that is not effective Pareto optimal has an alternative selection of arms over the same total number of rounds, resulting higher total rewards in all objectives.
    Our proposed algorithm, \texttt{MOL-TS}, operates on this new notion of the effective Pareto front and samples an arm from the estimated effective Pareto front.
    As a result, \texttt{MOL-TS} produces higher cumulative rewards compared to the methods that selected from the plain Pareto front.
    
    \item We establish that \texttt{MOL-TS} is statistically efficient, achieving the Pareto regret bound of $\widetilde O(d^{3/2}\sqrt{T})$, where $d$ is the dimension of the feature vectors, $T$ is the total number of rounds. 
    In order to ensure the provable guarantees of the randomized exploration for multiple objectives, \texttt{MOL-TS} adopts the \textit{optimistic sampling strategy} (Section~\ref{sec:opt-sampling}).

    \item Numerical experiments demonstrate the effectiveness of our proposed approach, showing improved performance in regret minimization, and objective-wise total reward maximization.
\end{itemize}

\section{Related works}

Multi-objective multi armed bandit setting was first explored by \citet{drugan2013momab}, who proposed UCB algorithms for MOMAB by applying two representative approaches: using Pareto order and scalarized order. Subsequently, ~\citet{auer2016pareto} proposed algorithms that identify all Pareto optimal arms with high probability.
More recently, the upper and lower bounds of Pareto regret in the MOMAB setting have been studied in both stochastic and adversarial settings by \citet{xu2023pareto}.
There are also several studies on multi-objective contextual bandits.
For example, \citet{tekin2018multi} studied MOMAB in a contextual setting where a dominant objective exists, but we do not assume any dominance among objectives.
\citet{turgay2018multi} developed the PCZ algorithm, which identifies the Pareto front using the idea of contextual zooming and proved its regret bound. However, the algorithm is complex, and the paper does not provide specific details on its implementation.
\citet{lu2019moglb} studied the multi-objective generalized linear bandit (MOGLB) problem and analyzed the upper bound of Pareto regret using the ParetoUCB algorithm.
Additionally, \citet{Kim2023} explored Pareto front identification in linear bandit settings. 
The studies mentioned thus far proposed complex algorithms that calculate the empirical Pareto front. 
In contrast, \citet{zhang2024} introduced a hypervolume scalarization method in stochastic linear bandit settings, which uses random scalarization to explore the entire Pareto front.

While these studies address significant challenges in multi-objective bandits, surprisingly, although the practical effectiveness of randomized methods is widely recognized, research on randomized algorithms in multi-objective bandits has been rare. To the best of our knowledge, only \citet{yahyaa2015thompson} proposed a Thompson Sampling (TS) algorithm for MOMAB, but no theoretical analysis of this approach has been conducted. 
Separately, there has been significant research on randomized scalarization in the multi-objective Bayesian optimization problem~\citep{paria2020flexible, zhang2020random}, including theoretical analyses of TS algorithms. 
However, \citet{zhang2020random} and  \citet{paria2020flexible} analyzed the "Bayes regret" with known Gaussian prior setting, increasing with the number of objectives.

In the single-objective case, the theoretical analysis of Thompson sampling was first introduced by ~\citet{agrawal2012analysis}, and then was extended to the stochastic linear bandit setting in ~\citet{agrawal2013thompson}, where arms were categorized as either saturated or unsaturated to derive theoretical bounds. 
From a different perspective, \citet{abeille2017linear} analyzed Thompson Sampling under the assumption of fixed probabilities for sampling optimistic parameters.
Additionally, \citet{chapelle2011empirical} showed that, although the theoretical guarantees of Thompson Sampling are weaker than those of UCB, empirical results have consistently demonstrated that TS algorithms outperform UCB algorithms in practice.
However, there has been a clear gap in extending these theoretical guarantees from single objective settings to multi-objective bandits.

We provide the first theoretical analysis of a randomized algorithm in the multi-objective bandit setting. To the best of our knowledge, this is the first work to propose a TS algorithm for multi-objective linear contextual bandits and to analyze it theoretically.

\section{Preliminaries} \label{sec:preliminaries}
\subsection{Notations}
Throughout this paper, we use notations that distinguish between different objectives. For any positive integer $N\in\NN$, let $[N] := \{1,2,\ldots,N\}$. We denote $L$ as the number of objectives, and for $\ell\in[L]$, any real number $u$ corresponding to the $\ell_{th}$-objective is denoted as $u\obj{\ell}$. The vector $\ub \in \RR^L$ comprises all $u\obj{\ell}$ values and is represented in bold notation, i.e., $ \ub = [
        u\obj{1} , u\obj{2} , \ldots , u\obj{L}
    ]^\top$.
Otherwise, individual features of any vector $x$ are typically denoted as $x(i)$. For clarity, $\|\cdot\|$ denotes the Euclidean norm, and for a positive semi-definite matrix $V$, the norm $\|\cdot\|_V$ is defined in the inner product space with the matrix $V$ as $\langle x,y\rangle_V = \sqrt{x^\top Vy}$. Finally, we define $\Scal^n\subset\RR^{n}$ as the unit $(n-1)$-simplex.

\subsection{Problem settings}
We consider a standard stochastic linear contextual bandit problem, extended to the multi-objective setting. Let $\Acal$ be a finite set of arms. Each arm $a \in \Acal$ corresponds to a $d$-dimensional context vector $x_{t,a}\in\RR^d$ which is adversarially given at each round $t$. For each objective $\ell\in[L]$, there exists a fixed parameter $\theta_*\obj{\ell}\in\RR^d$, but unknown to agent. In total, there are $L$ parameters $\theta_*\obj{1}, \theta_*\obj{2}, \ldots, \theta_*\obj{L}$. At each round $t\in [T]$, the agent selects an arm $a_t\in\Acal$ and receives a $L$-dimensional reward vector $\rb_{t,a_t} =[
    r_{t,a_t}\obj{1}, r_{t,a_t}\obj{2},  \ldots  ,r_{t,a_t}\obj{L}
    ]^\top \in \RR^L$,
where the reward for each objective $\ell$ is given by $r_{t,a_t}\obj{\ell} = x_{t,a_t}^\top \theta_*\obj{\ell} + \xi\obj{\ell}_t$, and $\xi\obj{\ell}_t$ is a zero-mean random noise. The mean reward for objective $\ell$ is defined as $\mu\obj{\ell}_{t,a_t} := \EE[r_{t,a_t}\obj{\ell}]$. And consequently, the mean reward vector of the chosen arm $a_t$ is $\mub_{t,a_t} = [
        \mu\obj{1}_{t,a_t} , \mu\obj{2}_{t,a_t} , \ldots , \mu\obj{L}_{t,a_t}
    ]^\top\in\RR^{L}$.

\begin{figure*}
\vskip 0.2in
\begin{center}
\centerline{\includegraphics[width=\linewidth]{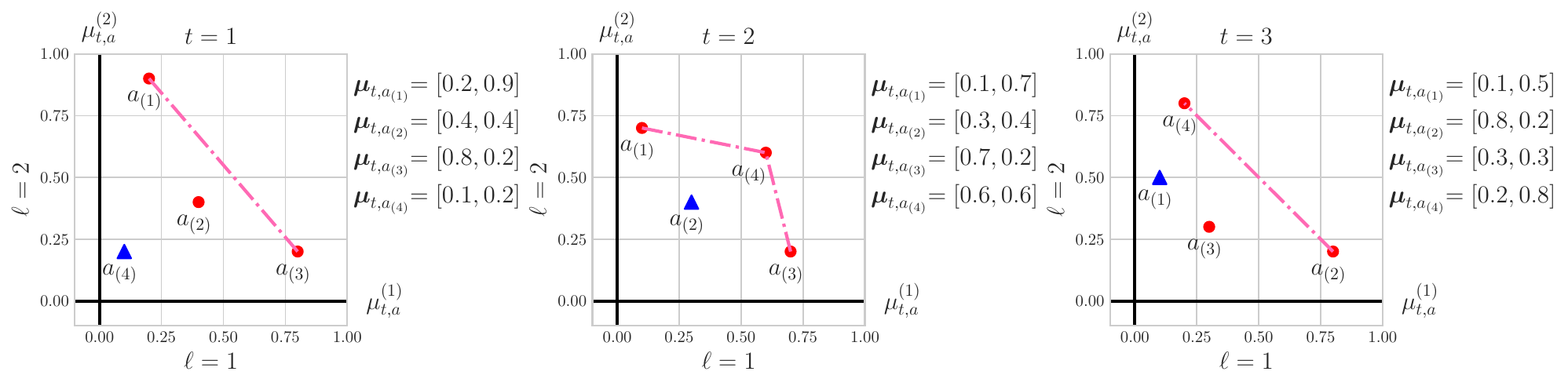}}
\caption{Example of two objectives and four arms, $a_{(1)}, a_{(2)}, a_{(3)}$, and $a_{(4)}$. Each subplot shows the mean reward vector at round $t$, where the horizontal and vertical axes correspond to the first and second objective, respectively. Red circles represent Pareto optimal arms, blue triangles that are not. The mean reward vectors are listed on the right, and the pink line represents the boundary of effective Pareto front (see \cref{def:CPO}).}
\label{fig:pareto-front}
\end{center}
\vskip -0.2in
\end{figure*}
\subsection{Pareto optimality} \label{section:p}

\begin{definition}[Pareto order] Let $\ub,\vb \in\mathbb{R}^L$ be two distinct vectors. We say that the vector $\ub$ is dominated by the vector $\vb$ (i.e., $\vb$ dominates $\ub$), denoted as $\ub\prec \vb$, if $u^{(\ell)}\le v^{(\ell)}$ for all $ \ell\in[L]$ and $u^{(\ell)}< v^{(\ell)}$ for some $\ell\in[L]$. Conversely, the vector $\ub$ is not dominated by the vector $\vb$, denoted as $\ub\not\prec \vb$, if there exists at least one $\ell\in[L]$ satisfying $u^{(\ell)}>v^{(\ell)}$
\end{definition} 
\begin{definition}[Pareto optimal arm]
    An arm is Pareto optimal if its mean reward vector is not dominated by that of any other arm. The set of all Pareto optimal arms is called Pareto Front $(\Pcal^*_t)$,  
    \begin{equation*}
    \Pcal^*_t := \{a\in\Acal\mid\mub_{t,a}\not\prec \mub_{t,a'},\forall a'\in\Acal\}.
\end{equation*}
In linear setting, where the mean rewards of each arm remain fixed, the Pareto front is denoted as $\Pcal^*$.
\end{definition}
\begin{definition}[Pareto sub-optimality gap]\label{def:psubgap}
    The Pareto sub-optimality gap of an arm $a$ at round $t$ is the minimum scalar value $\epsilon\ge0$ for $a$ to be Pareto optimal, i.e., 
    \begin{equation*}
        \Delta^{PR}_{t,a} := \inf\{\epsilon\ge0 \mid \mub_{t,a}+\epsilon \one\not\prec\mub_{t,a'}, \forall a'\in\Acal\}.
    \end{equation*}
\end{definition}
The Pareto sub-optimal gap can be defined as $\Delta^{PR}_{t,a} := \max_{a'\in\Pcal^*}\min_{\ell\in[L]}\{\mu\obj{\ell}_{t,a'}-\mu\obj{\ell}_{t,a}\}.$ 
For every Pareto optimal arm $a'\in\Pcal^*_t$, the arm that maximizes $\Delta^{PR}_{t,a'}$ is $a'$ itself, which implies $\Delta^{PR}_{t,a'} = 0$. Any other arm is dominated by at least one Pareto optimal arm, ensuring that $\Delta^{PR}_{t,a} \ge 0$.

\begin{definition}[Pareto regret]
    Let $a_1, a_2, \ldots, a_T$ be the sequence of arms chosen by agent. The Pareto regret up to round $T$ is defined as $PR(T):=\sum_{t=1}^T \Delta^{PR}_{t,a_t}$.
\end{definition}
In the contextual bandit setting, the Pareto front varies dynamically depending on the given context. Hence, we can not apply the algorithm of identifying Pareto front in \citet{auer2016pareto} and \citet{Kim2023} as they remove arm from the arm set, which can be Pareto optimal in our setting. 

By using the Pareto order relationship, the definition of Pareto regret provides a general measurement of an agent's performance in a multi-objective setting. 
Previous studies of Pareto optimality \cite{drugan2013momab,lu2019moglb, turgay2018multi, Kim2023} adopt this measurement and design algorithms that randomly select arms from the Pareto front, aiming to minimize the Pareto regret.
However, this definition of Pareto regret does not fully account for cumulative rewards. 
For example, consider a case with two objectives and four arms, as illustrated in \cref{fig:pareto-front}.
Suppose two agents follow same policy that randomly selects arm from Pareto front.
The first agent sequentially selects $a_{(2)}, a_{(4)}$ and $a_{(3)}$, while the second agent selects $a_{(1)}$, $a_{(4)}$ and $a_{(2)}$. 
Both agents selected Pareto optimal arms, resulting zero Pareto regret. 
But total rewards of the first agent is $\mub_{1,a_{(2)}} + \mub_{2,a_{(4)}} + \mub_{3,a_{(3)}} = [1.3, 1.3]$ and the second agent is $[1.6, 1.7]$. 
This example highlights the limitation of Pareto regret, as it does not distinguish between policies that yield different cumulative rewards despite selecting only Pareto optimal arms. 
Thereby, we propose the concept of a \textit{effective Pareto optimal} arm, whose mean reward vector is Pareto optimal and contributes to maximizing cumulative rewards across all objectives.

\subsection{Effective Pareto optimality}

\begin{definition}[Effective Pareto optimal arm]\label{def:CPO}
    An arm is effective Pareto optimal (denoted $a_*$) if its mean reward vector is either equal to or not dominated by any convex combination of the mean reward vectors of the other arms. Formally, for any $\beta\in\Scal^{|\Acal|-1}$, 
\begin{equation*}
    \mub_{t,a_*} = \sum_{a\in\Acal\setminus\{a_*\}}\beta_{a}\mub_{t,a}\qquad \text{ or } \qquad\mub_{t,a_*} \not\prec \sum_{a\in\Acal\setminus\{a_*\}}\beta_{a}\mub_{t,a},
\end{equation*} where $\beta = (\beta_a)_{a\in\Acal\setminus\{a_*\}}$.
The set of all effective Pareto optimal arms at round $t$ is called the effective Pareto front, denoted as $\Ccal_t^*$. In the linear bandit setting, where the mean reward vectors of all arms remain fixed, the effective Pareto front is denoted as $\Ccal^*$
\end{definition}
In this paper, we refer to an arm that is not effective Pareto optimal as sub-optimal. 
If an arm $a'\in\Acal$ is sub-optimal, then there exists a convex combination $\beta$, such that $\mub_{t,a'} \prec \sum_{a\in\Acal\setminus\{a'\}}\beta_{a}\mub_{t,a}.$

Every effective Pareto optimal arm is also a Pareto optimal arm. 
This can be easily verified by restricting $\beta$ to be a one-hot vector, which corresponds to comparing mean reward vectors between two individual arms. 
However, the converse does not hold. 
As can be seen from \cref{fig:pareto-front}, at the first round, arm $a_{(2)}$ is Pareto optimal, but not effective Pareto optimal, because its mean reward vector is dominated by a convex combination of two arms $a_{(1)}$ and $a_{(3)}$. 
Hence, for any $t\in[T]$, we have $\Ccal_t^* \subset \Pcal_t^*$. This shows that our definition of effective Pareto optimal arm is strictly defined than the standard definition of Pareto optimal arm.

As shown in the example in the previous subsection, two agents following the same policy randomly selected arms from the Pareto front. 
The second agent consistently selected arms from the effective Pareto front, while the first agent selected arms from the Pareto front but not the effective Pareto front in the first and third rounds. 
This difference led to the first agent achieving lower cumulative rewards than the second agent for all objectives. Importantly, this disparity becomes increasingly severe as the total number of rounds $T$ grows, leading to significantly worse long-term performance for policies that fail to prioritize the effective Pareto front.

The intuition behind an effective Pareto optimal is that repeatedly selecting such arms leads to Pareto optimal cumulative rewards.
In other words, rather than selecting a sub-optimal arm over multiple rounds, it is preferable to select effective Pareto optimal arms for the same total number of rounds, which is expected to yield strictly higher cumulative reward in some objectives without worsening the others.
In summary, for large enough total number of rounds $T$, selecting arms from effective Pareto front $\Ccal_t^*$ achieves higher total rewards than selecting arms from Pareto optimal front $\Pcal_t^*$. 
Based on this, we propose a theorem that establishes a relationship between the newly defined effective Pareto optimality and the linear scalarization method.
\begin{theorem} \label{thm:ptow}
    For any $a_*\in\Ccal^*$, there exist $\wb\in \Scal^L$ satisfying $a_* = \arg\max_{a\in\Acal} \wb^\top\mub_a$. Conversely, for any $\wb\in \Scal^L$, if $a_* = \arg\max_{a\in\Acal} \wb^\top\mub_a$ is unique arm, then $a_*\in\Ccal^*$.
\end{theorem}
The theorem is proved in \cref{app:A} where we refer to the proof from \citet{mangasarian1994nonlinear}. The theorem shows a one-to-one correspondence: every effective Pareto optimal arm is optimal for some non-negative weight vector, and every non-negative weight vector guarantees to have an effective Pareto optimal arm.

\begin{definition}[Effective Pareto sub-optimality gap] 
Let $\beta = (\beta_{a})_{a\in\Acal}$ be a vector in $\Scal^{|\Acal|}$ and define $\mub_{t,\beta} = \sum_{a\in\Acal}\beta_{a}\mub_{t,a}$. The effective Pareto sub-optimality gap for selecting arm $a_t$ at round $t$ is defined as 
\begin{align*}
    \Delta_{t,a_{t}}^{EPR} := \inf\Biggl\{\epsilon \ge 0 \mathrel{\bigg|} \mub_{t,a_t} +\epsilon \one \not\prec\mub_{t,\beta},\forall \beta\in\Scal^{|\Acal|}\Biggr\}.
\end{align*}
\end{definition}

The effective Pareto sub-optimal gap measures the minimum value $\epsilon$ for $a_t$ not to be dominated by any convex combination of the mean reward vectors of the other arms. In other words, the gap quantifies how close arm $a_t$ is to being effective Pareto optimal. For any effective Pareto optimal arm, this gap is zero.
Also, it is easy to verify that the effective Pareto sub-optimality gap is always greater than or equal to the standard Pareto sub-optimality gap, i.e., $\Delta_{t,a_t}^{PR} \le \Delta_{t,a_t}^{EPR}$, since the Pareto sub-optimality gap corresponds to the special case where $\beta$ is restricted to be a one-hot vector.
As discussed in \cref{section:p}, the effective Pareto sub-optimality gap can also be expressed as 
\begin{equation}\label{eq:EPR}
    \Delta_{t,a_{t}}^{EPR} := \max_{\beta\in\Scal^{|\Ccal_t^*|}}\min_{\ell\in[L]}\Biggl\{\biggl(\sum_{a_*\in\Ccal^*_t}\beta_{a_*}\mu_{t,a_*}\obj{\ell}\biggr)-\mu_{t,a_t}\obj{\ell}\Biggr\},
\end{equation}
\begin{definition}[Effective Pareto regret]
    The effective Pareto regret up to round $T$ is defined as $EPR(T) := \sum_{t = 1}^{T}\Delta_{t,a_t}^{EPR}$.
\end{definition}

\begin{algorithm}[tb]
    \caption{Multi-Objective Linear TS (\texttt{MOL-TS})} 
    \begin{algorithmic}
        \State {\bfseries Input:} $\lambda, M, T>0, c>0$
        \State {\bfseries Initialize:} $V_1 = \lambda I_{d},\,\hat\theta^{(\ell)}_1,Z_1\obj{\ell} =\zero\, (\forall \ell\in[L])$
        \For {$t = 1 \rightarrow T$}
            \For {objective $\ell = 1, 2, 3, \ldots, L$}
                \State Sample $ \tilde\theta_{t,1}\obj{\ell},\tilde\theta_{t,2}\obj{\ell},\ldots,\tilde\theta_{t,M}\obj{\ell} \sim \Ncal(\hat\theta_t\obj{\ell}, c^2V_t^{-1})$
                \State Evaluate every arm $\tilde \mu_{t,a}\obj{\ell}$ using \cref{eq:opt}
            \EndFor
            \State Update the empirical effective Pareto front $\tilde\Ccal_t$ using \cref{eq:eepf}
            \State Sample arm $a_t$ from $\tilde\Ccal_t$ uniformly at random, play $a_t$, observe reward vector $\rb_{t,a_t}$
            \State Update $V_{t+1}\leftarrow V_t + x_{t,a_t}x_{t,a_t}^\top$
            \For{objective $\ell = 1, \ldots, L$}
                \State Update $Z_{t+1}\obj{\ell} \leftarrow Z_{t}\obj{\ell} + x_{t,a_t}r_{t,a_t}\obj{\ell}$ and $\hat{\theta}\obj{\ell}_{t+1} \leftarrow V_{t+1}^{-1}Z_{t+1}\obj{\ell}$ 
            \EndFor
        \EndFor

    \end{algorithmic}
\end{algorithm}

\section{Algorithm: \texttt{MOL-TS}} \label{sec:algorithm}
We propose a multi-objective linear Thompson Sampling algorithm, \texttt{MOL-TS}, a generic randomized algorithm designed with multiple regularized MLE, where one need not sample from an actual Bayesian posterior ~\citep{abeille2017linear}. Our algorithm adopts an optimistic sampling strategy to avoid the theoretical challenges in worst-case regret analysis~\citep{oh2019advances, hwang2023combinatorial}. 

Each round $t$, the mean reward vector for each arm is estimated based on the history of chosen arms $a_1, a_2, \ldots, a_{t-1}$, and received reward vectors $\rb_{1,a_1}, \rb_{2,a_2}, \ldots, \rb_{t-1,a_{t-1}}$ up to round $t$. The true parameter for each objective $\theta_*^{(\ell)}$ is estimated by regularized least squares (RLS), denoted $\hat\theta_t^{(\ell)}$. Given regularizer $\lambda\in\mathbb{R}^+$, the matrix and the RLS estimator is defined as
\begin{equation*}\label{eq:estimate}
    V_t = \sum_{s=1}^{t-1}x_{s,a_s}x_{s,a_s}^\top + \lambda I_{d\times d},\qquad\hat\theta_t^{(\ell)} = V_t^{-1}\sum_{s=1}^{t-1}x_{s,a_s}r_{s,a_s}\obj{\ell}.
\end{equation*}
For each objective $\ell$, the parameters $(\tilde\theta_{t,m}\obj{\ell})_{m\in[M]}$ are sampled independently $M$ times from Gaussian posterior distribution $\Ncal(\hat\theta_t\obj{\ell}, c^2V_t^{-1})$, where the tunable parameters $c$ and $M$ are given from the beginning. A total of $ML$ samples are drawn.
The reward for each arm and for each objective is then optimistically evaluated using the sample that yields the highest value,
\begin{equation}\label{eq:opt}
    \tilde \mu\obj{\ell}_{t,a} = \max\{x_{t,a}^\top\tilde\theta\obj{\ell}_{t,1}, x_{t,a}^\top\tilde\theta\obj{\ell}_{t,2}, \ldots, x_{t,a}^\top\tilde\theta\obj{\ell}_{t,M}\}.
\end{equation}
The reward vector for each arm is then constructed as $ \tilde\mub_{t,a} = \begin{bmatrix}
        \tilde \mu\obj{1}_{t,a} & \tilde \mu\obj{2}_{t,a} & \ldots & \tilde \mu\obj{L}_{t,a}
    \end{bmatrix}^\top.$

The number of samples $M$ controls the probability that the estimated rewards are optimistically evaluated.  Increasing $M$ raises the likelihood that the reward estimates are optimistic, which is crucial for ensuring a high theoretical probability of optimism across multiple objectives (see \cref{sec:opt-sampling}). We approximate the empirical effective Pareto front $\tilde \Ccal_t$ using the estimated reward vectors, by
\begin{align}\label{eq:eepf}
    \tilde \Ccal_t = \Biggl\{ a\in\Acal \;\bigg|\; \tilde\mub_{t,a} = \sum_{a'\in\Acal}\beta_{a'}\tilde\mub_{t,a'} \text{ or }  \tilde\mub_{t,a} \not\prec \sum_{a'\in\Acal}\beta_{a'}\tilde\mub_{t,a'},\, \forall \beta_{a'}\in\Scal^{|\Acal|}\Biggr\}.
\end{align}
Note that this optimistic sampling strategy is different from that proposed in \citep{oh2019advances, hwang2023combinatorial}. The setting in \citep{oh2019advances} considers a dynamic assortment selection problem, and \citep{hwang2023combinatorial} considers a combinatorial selection problem. Unlike multiple arms selection problem in single objective setting, our setting considers receiving multiple rewards from single arm selection problem.

\section{Regret analysis} \label{sec:analysis}

In this section, we analyze the expected effective Pareto regret of our algorithm \texttt{MOL-TS} in the worst-case, where the expectation is taken over all sources of randomness present in the problem setup. We begin with the general assumptions widely used in the linear bandit literature. We then outline the challenges in bounding the effective Pareto regret and explain how the number of samples $M$ affects the worst-case regret bound.

\subsection{Assumptions}

Let $\Fcal_t = \sigma(x_{1,a_1},\ldots,x_{t,a_t},\rb_{1,a_1},\ldots,\rb_{t-1,a_{t-1}})$ be the filtration up to round $t$ containing all historical information about the selected arms and the received rewards. The following assumptions are commonly used in the stochastic linear bandit literature. 
\begin{assumption}[Boundedness] \label{asump:1}
    For each arm $a\in\Acal$, $\|x_{t,a}\|\le1$.  Also, $\|\theta_*\obj{\ell}\|\le 1$ for all $\ell\in[L]$.
\end{assumption}

\begin{assumption}[Sub-Gaussian] \label{asump:2}
    Each noise $\xi\obj{\ell}_t$ is conditionally $R$-sub-Gaussian, given the filtration $\Fcal_t$ and for some $R\in\RR^+$.
\end{assumption}

Under the first assumption, each vector is both fixed and bounded for all rounds. If $\|x_{t,a}\|\le C$ and $\|\theta_*\obj{\ell}\|\le C$ are bounded for some constant $C$, then our regret bound would increase by a factor of $C$.
Note that we do not assume any linear independence of the true parameters or noise vectors between objectives. Our assumptions are essentially the same as those used in the single objective stochastic linear bandit setting.

\subsection{Challenges in bounding the effective Pareto regret}\label{sec:bound}
Previously, there are many papers of multi-objective UCB-type algorithm with Pareto regret analysis \cite{drugan2013momab,tekin2018multi,turgay2018multi,lu2019moglb,Kim2023,xu2023pareto}. The analysis of UCB algorithm is almost similar to that of single objective setting, attaining Pareto regret bound
where the number of objectives depend up to logarithmic factor. By contrast, deriving comparable guarantees for TS in the multi-objective linear contextual setting remains an open problem, and our work is the first to tackle these technical obstacles directly.

Recall the effective Pareto regret \cref{eq:EPR}. For any weight vector $\wb\in \Scal^L$, since $\|\wb\|_1 = 1$ ,
\begin{equation*}
    \min_{\ell\in[L]}\Biggl\{\biggl(\sum_{a_*\in\Ccal_t^*}\beta_{a_*}\mu_{t,a_*}\obj{\ell}\biggr)-\mu_{t,a_t}\obj{\ell}\Biggr\}
    \le \wb^\top \Biggl(\biggl(\sum_{a_*\in\Ccal_t^*}\beta_{a_*}\mub_{t,a_*}\biggr)-\mub_{t,a_t}\Biggr).
\end{equation*}

The algorithm \texttt{MOL-TS} optimistically evaluates reward vector $\tilde\mub_{t,a}$ for each arm, and selects arm $a_t$ randomly from the set $\tilde \Ccal_t$. 
Hence, by \cref{thm:ptow}, there exist weight vector, denoted $\wb_{t}$, satisfying $a_t=\arg\max_{a\in\Acal} \wb_{t}^\top\tilde\mub_{t,a}$. 
But for true mean reward vector, let $\bar a_*=\arg\max_{a\in\Acal} \wb_{t}^\top\mub_{t,a}$ be the effective Pareto optimal arm for given weight vector $\wb_t$.
Then we have
\begin{equation*}
    \Delta_{t,a_{t}}^{EPR}\le \max_{\beta \in \Scal^{|\Ccal_t^*|}}\Biggl\{\wb_t^\top \Biggl(\biggl(\sum_{a_*\in\Ccal^*}\beta_{a_*}\mub_{t,a_*}\biggr)-\mub_{t,a_t}\Biggr)\Biggr\}
    \le \wb_t^\top(\mub_{t,\bar a_*} - \mub_{t,a_t}).
\end{equation*}

The key insight is that the effective Pareto regret is bounded by the weighted sum of rewards under an arbitrary weight vector, and the same holds for Pareto regret. 
This analysis generalizes the single objective case, which is recovered by restricting $\wb_t$ to a one-hot vector.

However, since the arm $a_t$ is randomly selected from the set $\tilde \Ccal_t$, both the weight vector $\wb_t$, and the corresponding effective Pareto optimal arm $\bar a_*$ are random. 
Due to the randomness of the vector $\wb_t$ and the optimal arm $\bar a_*$, analyzing the worst-case regret bound of TS algorithm becomes significantly more difficult. Also, unlike the single objective setting, the multi-objective setting involves multiple true parameters and corresponding RLS estimates. This complicates the problem of ensuring optimism, as there are multiple sampled parameters in TS algorithm (see example in \cref{sec:opt-sampling}).
We resolve these theoretical challenges by adopting an \textit{optimistic sampling strategy}.

\subsection{Why do we need optimistic sampling?}\label{sec:opt-sampling}

In this section, we explain the necessity of the number of samples $M$. As we discuss in \cref{sec:bound}, the challenges in the worst-case regret analysis for TS algorithms lie in the difficulty of ensuring optimism in randomly selected arm $a_t$. When $\wb_t$ is one-hot vector, the analysis aligns with the single objective setting \cite{agrawal2013thompson, abeille2017linear}.
However, since $\wb_t$ is random, the analysis requires that the randomly chosen arm is optimistically evaluated under a weighted sum of rewards.
This probability can become exponentially small as the number of objectives increases.

Before providing a detailed explanation, we first define the event $\hat\Ecal_t$ that the true parameters $\theta_*\obj{\ell}$ are close enough to the RLS estimate parameters $\hat\theta_t\obj{\ell}$, and define $c_{1,t}(\delta)$ which is the high probability bound on the distance between the true parameter and the RLS estimate,
\begin{align*}
    \hat\Ecal_t :=\{\forall \ell \in[L] : \|\theta_*\obj{\ell}-\hat\theta_t\obj{\ell}\|_{V_t}\le c_{1,t}(\delta)\},\quad c_{1,t}(\delta) := R\sqrt{d\log\left(\frac{1+(t-1)/(\lambda d)}{\delta/L}\right)}+\lambda^{1/2}.
\end{align*}

By \cref{lem:abbasi1}, we have $\PP(\hat\Ecal_t) \ge 1-\delta$. The $O(\sqrt{\log L})$ dependence in $c_{1,t}(\delta)$ is inevitable as the confidence bound must hold uniformly over all objectives. We define the event $\dot\Ecal_{t,a}\obj{\ell}$ for a specific arm $a$ and objective $\ell$, where the event has at least one parameter following anti-concentration property of being optimistic, i.e.,
\begin{equation*}
     \dot\Ecal_{t,a}\obj{\ell} := \{ \exists m\in[M]: x_{t,a}^\top(\tilde\theta_{t,m}\obj{\ell}-\hat\theta_{t}\obj{\ell})\ge c_{1,t}(\delta)\|x_{t,a}\|_{V_t^{-1}}\}.
\end{equation*}

As the algorithm \texttt{MOL-TS} optimistically evaluate each arm $a$ with \cref{eq:opt}, the probability $\PP(\dot\Ecal_{t,a}\obj{\ell})$ increases as $M$ increases.
Suppose, for example, one follows standard TS algorithm by setting $M=1$, that only one parameter is sampled for each objective. Previous studies \cite{agrawal2013thompson, abeille2017linear} have shown that the probability of an arm $a$ being optimistically evaluated is at least $\tilde p$, i.e.,
\begin{equation*}
    \PP\{x_{t,a}^\top(\tilde\theta_{t,1}\obj{\ell}-\hat\theta_{t}\obj{\ell})\ge c_{1,t}(\delta)\|x_{t,a}\|_{V_t^{-1}}\} \ge \tilde p.
\end{equation*}  
where $\tilde p$ is constant probability, that depends on the choice of sampling distribution.
However, since $\wb_t$ is random, the probability of ensuring this optimism for every objective is at least $\tilde p^L$. Since this probability decreases exponentially with the number of objectives, the regret grows exponentially in $L$, yielding $\widetilde O(\nicefrac{1}{\tilde p^L}\cdot d^{3/2}\sqrt{T})$. 

Optimistic sampling strategy resolves this problem as \texttt{MOL-TS} optimistically evaluate the rewards using $M$ independent parameter samples for each objective. Specifically, the algorithm evaluates the arm according to the sampled parameter that maximizes the evaluation, i.e., \begin{equation*}
    \tilde\theta_{a,t}\obj{\ell} = \argmax_{(\tilde\theta_{t,m})_{m\in[M]}}\{x_{t,a}^\top\tilde\theta_{t,1}\obj{\ell}, x_{t,a}^\top\tilde\theta_{t,2}\obj{\ell}, ..., x_{t,a}^\top\tilde\theta_{t,M}\obj{\ell}\}.
\end{equation*}
Then, the probability bound for the optimism event is 
\begin{equation*}
    \PP\{x_{t,a}^\top(\tilde\theta_{a,t}\obj{\ell}-\hat\theta_{t}\obj{\ell})\ge c_{1,t}(\delta)\|x_{t,a}\|_{V_t^{-1}}\} \ge (1-(1-\tilde p)^M)^L.
\end{equation*}  
To prevent the exponential growth of the probability of ensuring optimism, the number of samples $M$ must depend on the number of objectives $L$. The next lemma shows the minimum number of samples $M$ for ensuring the event of optimism with constant probability.

\begin{lemma}[Optimistic Sampling]\label{lem:opt}
    For any arm $a\in\Acal$, define the event of anti-concentration property of being optimism $\dot\Ecal_{t,a} = \bigcap_{\ell\in[L]} \dot\Ecal_{t,a}\obj{\ell}$.
    Then on event $\hat\Ecal_t$, with $p = 0.15$ and $M \ge 1- \frac{\log L}{\log (1-p)}$, we have $\PP(\dot\Ecal_{t,a}) \ge p$.
\end{lemma}

The event $\dot\Ecal_{t,a}$ is that the arm $a$ is optimistically evaluated for every objective. \cref{lem:opt} shows that the lower bound on the probability that arm $a$ being optimistically evaluated remains constant by taking optimistic sampling strategy. The proof of this lemma is provided in \cref{prf:opt}.

\subsection{Worst-case regret bound}

We now present the worst-case (frequentist) regret upper bound of \texttt{MOL-TS}, where the expectation is taken over all sources of randomness present in the problem setup.

\begin{theorem}[Effective Pareto regret of \texttt{MOL-TS}] \label{thm:regret} 
With Assumptions \ref{asump:1} and \ref{asump:2}, with $c = c_{1,t}(\delta)$ and $M =\lceil 1- \frac{\log L}{\log (1-p)} \rceil$, the effective Pareto regret of the algorithm {\normalfont{\texttt{MOL-TS}}} is upper-bounded by 
    \begin{equation*}
        \EE [EPR(T)] = \left(1+\frac{2}{p-\frac{\delta}{T}}\right)c_T(\delta)\sqrt{2Td\log\left(1+\frac{T}{\lambda}\right)} + 2\delta \Delta_{\text{max}},
    \end{equation*}
where 
\begin{equation*}
    c_T(\delta) = \left(R\sqrt{d\log\left(\frac{1+(T-1)/(\lambda d)}{\delta/L}\right)}+\lambda^{1/2}\right)\left(1+\sqrt{2d\log\frac{2LMdT}{\delta}}\right).
\end{equation*}
\end{theorem}

\begin{corollary}[Pareto regret of \texttt{MOL-TS}]\label{cor:regret} 
With same assumptions and initialization, the Pareto regret of the algorithm {\normalfont{\texttt{MOL-TS}}} is upper-bounded by 
\begin{equation*}
    \EE [PR(T)] = \left(1+\frac{2}{p-\frac{\delta}{T}}\right)c_T(\delta)\sqrt{2Td\log\left(1+\frac{T}{\lambda}\right)} + 2\delta \Delta_{\text{max}}.
\end{equation*}
\end{corollary}

\textbf{Discussions of \cref{thm:regret} and \cref{cor:regret}.}
\cref{thm:regret} establishes that the expected effective Pareto regret of \texttt{MOL-TS} is bounded above by $\widetilde O(d^{3/2}\sqrt{T})$, where the regret has an additional $O(\log L)$ and $O(\sqrt{\log M})$ dependence on the number of objectives and the number samples, respectively, both of which are minimal. Additionally, \cref{cor:regret} holds since $\Delta_{t,a_t}^{PR} \le \Delta_{t,a_t}^{EPR}$. The details of the proof are provided in \cref{app:B}. To the best of our knowledge, \texttt{MOL-TS} is the first TS algorithm with the worst-case regret guarantees in both Pareto regret and effective Pareto regret.

\begin{figure}[tb]
    \centering
    \includegraphics[width=1.0\textwidth]{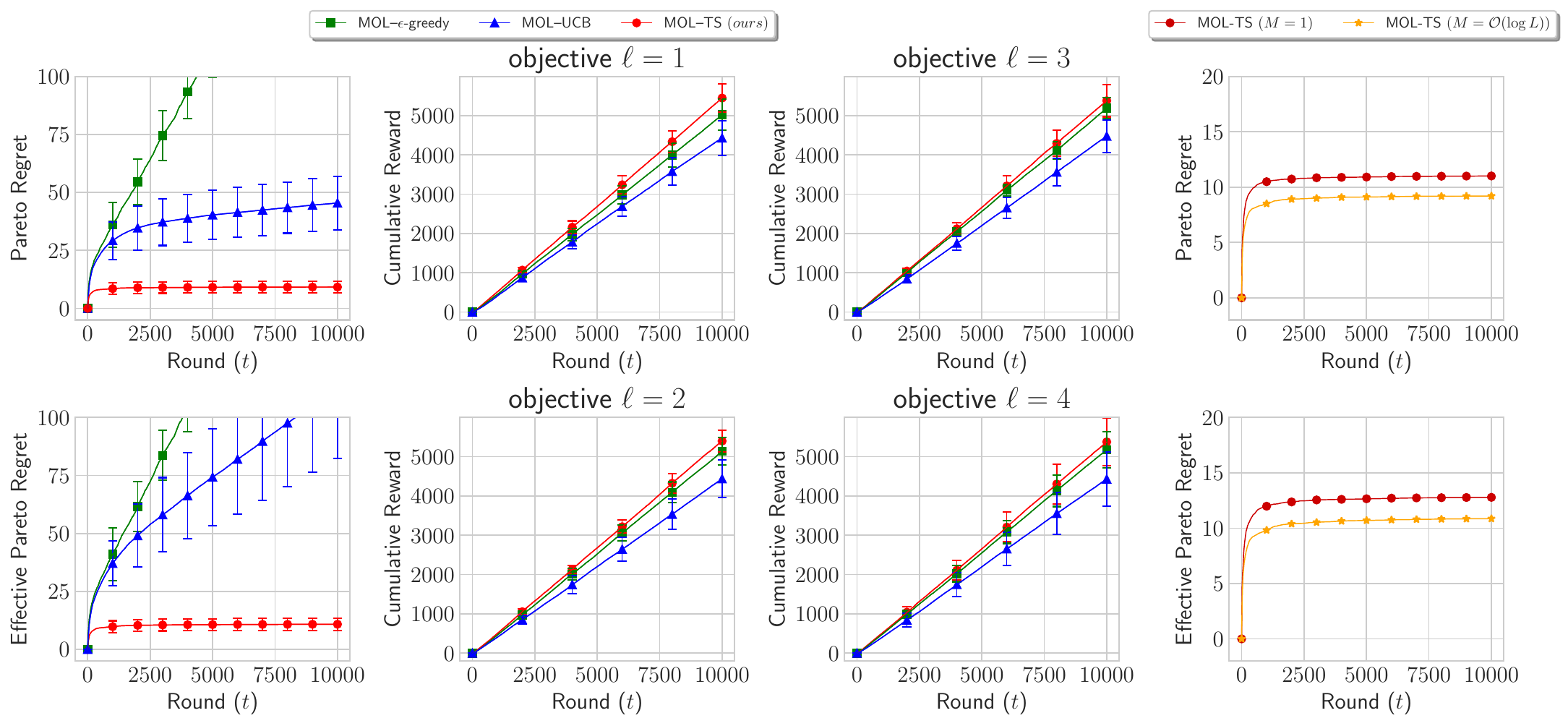}
    \vspace{-0.3cm}
     \caption{Experimental results with $4$ objectives ($L=4$). Plots in the left three columns measure the performances of \texttt{MOL-TS} and the others. Two plots in the first column measure the Pareto regret and the effective Pareto regret. Four plots in the second and third columns measure the cumulative reward for each objective. Plots in the right most column measure the performances of \texttt{MOL-TS} with $M = 1$ and $M = O(\log L)$. The error bars represent the 1-sigma standard deviation over $10$ instances.} 
    \label{fig:plot}
\end{figure}

\section{Experiments}\label{sec:experiment}

In this section, we empirically evaluate the performance of our algorithm. We measure the Pareto regret and effective Pareto regret over $T = 10000$ rounds. Each experimental setup contains 10 different instances with fixed number of arms $K$, objectives $L$, and feature dimension $d$. We demonstrate the case where $K = 50$, $d = 5$, $L = 4$. The parameter vector for each objective $\theta\obj{\ell}_*$ has a norm of $1$. Each round, $d$-dimensional context vectors are revealed for every arm, bounded by $1$ in Euclidean norm. Upon playing an arm, the agent receives a reward vector with an additional noise term, where the noise values are sampled from a zero mean Gaussian distribution with $\sigma = 1$.

We compare the performance of \texttt{MOL-TS} with those basic novel algorithms : the Upper Confidence Bound algorithm, and $\epsilon$-Greedy algorithm. The Upper Confidence Bound algorithm is \texttt{MOGLM-UCB} (represented as \texttt{MOL-UCB} in our experiments), from \citet{lu2019moglb} in linear bandit setting and $\epsilon$-Greedy algorithm \texttt{MOL-$\epsilon$-Greedy} is basic MOMAB algorithm with $\epsilon = 0.05$. Other algorithms cannot be applied in contextual setting, as they remove sub-optimal arm from the arm set. We also compare the performance of \texttt{MOL-TS} with and without the optimistic sampling. 
As shown in \cref{fig:plot}, our proposed algorithm \texttt{MOL-TS} shows greater performance compared to other algorithms, with minimizing the Pareto regret and effective Pareto regret, and maximizing cumulative rewards in all objectives. Additionally, \texttt{MOL-TS} with optimistic sampling performs better in minimizing regret. Additional experiments with different settings of $K, d$ and $L$ are left in \cref{app:result}, with additional algorithm \texttt{PFIwR} from \citet{Kim2023} in linear bandit setting.

We observe that \texttt{MOL-$\epsilon$-Greedy} yields higher Pareto regret and effective Pareto regret, but also total rewards compared to \texttt{MOL-UCB}. This counterintuitive behavior arises from the averaging of cumulative rewards: although the algorithm selects arms that are Pareto optimal, averaging their outcomes can reduce the overall performance because the algorithm randomly samples from the Pareto front. In other words, by exploring multiple Pareto optimal arms without a consistent preference direction, averaging total rewards may appear smaller despite balanced trade-offs. This issue could be mitigated by guiding the arm selection toward a specific scalarization or optimization direction, allowing the algorithm to maintain both Pareto efficiency and higher total reward.

\section{Discussions}
In this paper, we study the multi-objective linear contextual bandit problem, where multiple conflicting objectives must be optimized simultaneously. We define the effective Pareto regret, whose definition considers the Pareto optimality of cumulative reward vectors. We propose a Thompson Sampling algorithm with optimistic sampling strategy, \texttt{MOL-TS}, that achieves the Pareto regret and effective Pareto regret of $\widetilde O(d^{3/2}\sqrt{T})$, matching the best known order for randomized linear bandit algorithms for single objective setting. Empirical results confirm the benefits of our proposed approach, demonstrating improved regret minimization and strong multi-objective performance.



\section*{Acknowledgements}
This work was supported by the National Research Foundation of Korea~(NRF) grant funded by the Korea government~(MSIT) (No.  RS-2022-NR071853 and RS-2023-00222663), by Institute of Information \& communications Technology Planning \& Evaluation~(IITP) grant funded by the Korea government~(MSIT) (No. RS-2025-02263754), and by AI-Bio Research Grant through Seoul National University.

\bibliography{example_paper}
\bibliographystyle{plainnat}


\newpage
\section*{NeurIPS Paper Checklist}

\begin{enumerate}

\item {\bf Claims}
    \item[] Question: Do the main claims made in the abstract and introduction accurately reflect the paper's contributions and scope?
    \item[] Answer: \answerYes{} 
    \item[] Justification: We propose the first Thompson Sampling algorithm with Pareto regret guarantees in multi-objective linear contextual bandit. Our contributions are clearly summarized in \cref{sec:introduction}.
    \item[] Guidelines:
    \begin{itemize}
        \item The answer NA means that the abstract and introduction do not include the claims made in the paper.
        \item The abstract and/or introduction should clearly state the claims made, including the contributions made in the paper and important assumptions and limitations. A No or NA answer to this question will not be perceived well by the reviewers. 
        \item The claims made should match theoretical and experimental results, and reflect how much the results can be expected to generalize to other settings. 
        \item It is fine to include aspirational goals as motivation as long as it is clear that these goals are not attained by the paper. 
    \end{itemize}

\item {\bf Limitations}
    \item[] Question: Does the paper discuss the limitations of the work performed by the authors?
    \item[] Answer: \answerYes{} 
    \item[] Justification: We discuss the limitations of our work in \cref{app:limitation}.
    \item[] Guidelines:
    \begin{itemize}
        \item The answer NA means that the paper has no limitation while the answer No means that the paper has limitations, but those are not discussed in the paper. 
        \item The authors are encouraged to create a separate "Limitations" section in their paper.
        \item The paper should point out any strong assumptions and how robust the results are to violations of these assumptions (e.g., independence assumptions, noiseless settings, model well-specification, asymptotic approximations only holding locally). The authors should reflect on how these assumptions might be violated in practice and what the implications would be.
        \item The authors should reflect on the scope of the claims made, e.g., if the approach was only tested on a few datasets or with a few runs. In general, empirical results often depend on implicit assumptions, which should be articulated.
        \item The authors should reflect on the factors that influence the performance of the approach. For example, a facial recognition algorithm may perform poorly when image resolution is low or images are taken in low lighting. Or a speech-to-text system might not be used reliably to provide closed captions for online lectures because it fails to handle technical jargon.
        \item The authors should discuss the computational efficiency of the proposed algorithms and how they scale with dataset size.
        \item If applicable, the authors should discuss possible limitations of their approach to address problems of privacy and fairness.
        \item While the authors might fear that complete honesty about limitations might be used by reviewers as grounds for rejection, a worse outcome might be that reviewers discover limitations that aren't acknowledged in the paper. The authors should use their best judgment and recognize that individual actions in favor of transparency play an important role in developing norms that preserve the integrity of the community. Reviewers will be specifically instructed to not penalize honesty concerning limitations.
    \end{itemize}

\item {\bf Theory assumptions and proofs}
    \item[] Question: For each theoretical result, does the paper provide the full set of assumptions and a complete (and correct) proof?
    \item[] Answer: \answerYes{} 
    \item[] Justification: Theorems are presented with assumptions in detail (see \cref{sec:analysis}). Details and explanations of the proofs are given in \cref{app:A} and \cref{app:B}.
    \item[] Guidelines:
    \begin{itemize}
        \item The answer NA means that the paper does not include theoretical results. 
        \item All the theorems, formulas, and proofs in the paper should be numbered and cross-referenced.
        \item All assumptions should be clearly stated or referenced in the statement of any theorems.
        \item The proofs can either appear in the main paper or the supplemental material, but if they appear in the supplemental material, the authors are encouraged to provide a short proof sketch to provide intuition. 
        \item Inversely, any informal proof provided in the core of the paper should be complemented by formal proofs provided in appendix or supplemental material.
        \item Theorems and Lemmas that the proof relies upon should be properly referenced. 
    \end{itemize}

    \item {\bf Experimental result reproducibility}
    \item[] Question: Does the paper fully disclose all the information needed to reproduce the main experimental results of the paper to the extent that it affects the main claims and/or conclusions of the paper (regardless of whether the code and data are provided or not)?
    \item[] Answer: \answerYes{} 
    \item[] Justification: All codes of algorithms and experiments are provided in a ZIP file. Experimental results are provided in \cref{sec:experiment} and \cref{app:result}.
    \item[] Guidelines:
    \begin{itemize}
        \item The answer NA means that the paper does not include experiments.
        \item If the paper includes experiments, a No answer to this question will not be perceived well by the reviewers: Making the paper reproducible is important, regardless of whether the code and data are provided or not.
        \item If the contribution is a dataset and/or model, the authors should describe the steps taken to make their results reproducible or verifiable. 
        \item Depending on the contribution, reproducibility can be accomplished in various ways. For example, if the contribution is a novel architecture, describing the architecture fully might suffice, or if the contribution is a specific model and empirical evaluation, it may be necessary to either make it possible for others to replicate the model with the same dataset, or provide access to the model. In general. releasing code and data is often one good way to accomplish this, but reproducibility can also be provided via detailed instructions for how to replicate the results, access to a hosted model (e.g., in the case of a large language model), releasing of a model checkpoint, or other means that are appropriate to the research performed.
        \item While NeurIPS does not require releasing code, the conference does require all submissions to provide some reasonable avenue for reproducibility, which may depend on the nature of the contribution. For example
        \begin{enumerate}
            \item If the contribution is primarily a new algorithm, the paper should make it clear how to reproduce that algorithm.
            \item If the contribution is primarily a new model architecture, the paper should describe the architecture clearly and fully.
            \item If the contribution is a new model (e.g., a large language model), then there should either be a way to access this model for reproducing the results or a way to reproduce the model (e.g., with an open-source dataset or instructions for how to construct the dataset).
            \item We recognize that reproducibility may be tricky in some cases, in which case authors are welcome to describe the particular way they provide for reproducibility. In the case of closed-source models, it may be that access to the model is limited in some way (e.g., to registered users), but it should be possible for other researchers to have some path to reproducing or verifying the results.
        \end{enumerate}
    \end{itemize}

\item {\bf Open access to data and code}
    \item[] Question: Does the paper provide open access to the data and code, with sufficient instructions to faithfully reproduce the main experimental results, as described in supplemental material?
    \item[] Answer: \answerYes{} 
    \item[] Justification: All codes of algorithms and experiments are provided in a ZIP file. Experimental results are provided in \cref{sec:experiment} and \cref{app:result}.
    \item[] Guidelines:
    \begin{itemize}
        \item The answer NA means that paper does not include experiments requiring code.
        \item Please see the NeurIPS code and data submission guidelines (\url{https://nips.cc/public/guides/CodeSubmissionPolicy}) for more details.
        \item While we encourage the release of code and data, we understand that this might not be possible, so “No” is an acceptable answer. Papers cannot be rejected simply for not including code, unless this is central to the contribution (e.g., for a new open-source benchmark).
        \item The instructions should contain the exact command and environment needed to run to reproduce the results. See the NeurIPS code and data submission guidelines (\url{https://nips.cc/public/guides/CodeSubmissionPolicy}) for more details.
        \item The authors should provide instructions on data access and preparation, including how to access the raw data, preprocessed data, intermediate data, and generated data, etc.
        \item The authors should provide scripts to reproduce all experimental results for the new proposed method and baselines. If only a subset of experiments are reproducible, they should state which ones are omitted from the script and why.
        \item At submission time, to preserve anonymity, the authors should release anonymized versions (if applicable).
        \item Providing as much information as possible in supplemental material (appended to the paper) is recommended, but including URLs to data and code is permitted.
    \end{itemize}

\item {\bf Experimental setting/details}
    \item[] Question: Does the paper specify all the training and test details (e.g., data splits, hyperparameters, how they were chosen, type of optimizer, etc.) necessary to understand the results?
    \item[] Answer: \answerYes{} 
    \item[] Justification: The details of experimental settings are provided in \cref{sec:experiment}.
    \item[] Guidelines:
    \begin{itemize}
        \item The answer NA means that the paper does not include experiments.
        \item The experimental setting should be presented in the core of the paper to a level of detail that is necessary to appreciate the results and make sense of them.
        \item The full details can be provided either with the code, in appendix, or as supplemental material.
    \end{itemize}

\item {\bf Experiment statistical significance}
    \item[] Question: Does the paper report error bars suitably and correctly defined or other appropriate information about the statistical significance of the experiments?
    \item[] Answer: \answerYes{} 
    \item[] Justification: In both \cref{sec:experiment} and \cref{app:result}, all the bars in regret plots and reward plots represent the $1$-sigma standard deviation of experimental results.
    \item[] Guidelines: 
    \begin{itemize}
        \item The answer NA means that the paper does not include experiments.
        \item The authors should answer "Yes" if the results are accompanied by error bars, confidence intervals, or statistical significance tests, at least for the experiments that support the main claims of the paper.
        \item The factors of variability that the error bars are capturing should be clearly stated (for example, train/test split, initialization, random drawing of some parameter, or overall run with given experimental conditions).
        \item The method for calculating the error bars should be explained (closed form formula, call to a library function, bootstrap, etc.)
        \item The assumptions made should be given (e.g., Normally distributed errors).
        \item It should be clear whether the error bar is the standard deviation or the standard error of the mean.
        \item It is OK to report 1-sigma error bars, but one should state it. The authors should preferably report a 2-sigma error bar than state that they have a 96\% CI, if the hypothesis of Normality of errors is not verified.
        \item For asymmetric distributions, the authors should be careful not to show in tables or figures symmetric error bars that would yield results that are out of range (e.g. negative error rates).
        \item If error bars are reported in tables or plots, The authors should explain in the text how they were calculated and reference the corresponding figures or tables in the text.
    \end{itemize}

\item {\bf Experiments compute resources}
    \item[] Question: For each experiment, does the paper provide sufficient information on the computer resources (type of compute workers, memory, time of execution) needed to reproduce the experiments?
    \item[] Answer: \answerYes{} 
    \item[] Justification: We include information about the computing environment used to run experiments in \cref{app:resources}.
    \item[] Guidelines:
    \begin{itemize}
        \item The answer NA means that the paper does not include experiments.
        \item The paper should indicate the type of compute workers CPU or GPU, internal cluster, or cloud provider, including relevant memory and storage.
        \item The paper should provide the amount of compute required for each of the individual experimental runs as well as estimate the total compute. 
        \item The paper should disclose whether the full research project required more compute than the experiments reported in the paper (e.g., preliminary or failed experiments that didn't make it into the paper). 
    \end{itemize}
    
\item {\bf Code of ethics}
    \item[] Question: Does the research conducted in the paper conform, in every respect, with the NeurIPS Code of Ethics \url{https://neurips.cc/public/EthicsGuidelines}?
    \item[] Answer: \answerYes{} 
    \item[] Justification: The research in our paper conforms with the NeurIPS Code of Ethics.
    \item[] Guidelines:
    \begin{itemize}
        \item The answer NA means that the authors have not reviewed the NeurIPS Code of Ethics.
        \item If the authors answer No, they should explain the special circumstances that require a deviation from the Code of Ethics.
        \item The authors should make sure to preserve anonymity (e.g., if there is a special consideration due to laws or regulations in their jurisdiction).
    \end{itemize}

\item {\bf Broader impacts}
    \item[] Question: Does the paper discuss both potential positive societal impacts and negative societal impacts of the work performed?
    \item[] Answer: \answerNA{} 
    \item[] Justification: Our paper has no societal impact of the work performed.
    \item[] Guidelines:
    \begin{itemize}
        \item The answer NA means that there is no societal impact of the work performed.
        \item If the authors answer NA or No, they should explain why their work has no societal impact or why the paper does not address societal impact.
        \item Examples of negative societal impacts include potential malicious or unintended uses (e.g., disinformation, generating fake profiles, surveillance), fairness considerations (e.g., deployment of technologies that could make decisions that unfairly impact specific groups), privacy considerations, and security considerations.
        \item The conference expects that many papers will be foundational research and not tied to particular applications, let alone deployments. However, if there is a direct path to any negative applications, the authors should point it out. For example, it is legitimate to point out that an improvement in the quality of generative models could be used to generate deepfakes for disinformation. On the other hand, it is not needed to point out that a generic algorithm for optimizing neural networks could enable people to train models that generate Deepfakes faster.
        \item The authors should consider possible harms that could arise when the technology is being used as intended and functioning correctly, harms that could arise when the technology is being used as intended but gives incorrect results, and harms following from (intentional or unintentional) misuse of the technology.
        \item If there are negative societal impacts, the authors could also discuss possible mitigation strategies (e.g., gated release of models, providing defenses in addition to attacks, mechanisms for monitoring misuse, mechanisms to monitor how a system learns from feedback over time, improving the efficiency and accessibility of ML).
    \end{itemize}
    
\item {\bf Safeguards}
    \item[] Question: Does the paper describe safeguards that have been put in place for responsible release of data or models that have a high risk for misuse (e.g., pretrained language models, image generators, or scraped datasets)?
    \item[] Answer: \answerNA{} 
    \item[] Justification: Our paper poses no such risks.
    \item[] Guidelines:
    \begin{itemize}
        \item The answer NA means that the paper poses no such risks.
        \item Released models that have a high risk for misuse or dual-use should be released with necessary safeguards to allow for controlled use of the model, for example by requiring that users adhere to usage guidelines or restrictions to access the model or implementing safety filters. 
        \item Datasets that have been scraped from the Internet could pose safety risks. The authors should describe how they avoided releasing unsafe images.
        \item We recognize that providing effective safeguards is challenging, and many papers do not require this, but we encourage authors to take this into account and make a best faith effort.
    \end{itemize}

\item {\bf Licenses for existing assets}
    \item[] Question: Are the creators or original owners of assets (e.g., code, data, models), used in the paper, properly credited and are the license and terms of use explicitly mentioned and properly respected?
    \item[] Answer: \answerYes{} 
    \item[] Justification: We clearly mention the sources of the comparator algorithms in \cref{sec:experiment}.
    \item[] Guidelines:
    \begin{itemize}
        \item The answer NA means that the paper does not use existing assets.
        \item The authors should cite the original paper that produced the code package or dataset.
        \item The authors should state which version of the asset is used and, if possible, include a URL.
        \item The name of the license (e.g., CC-BY 4.0) should be included for each asset.
        \item For scraped data from a particular source (e.g., website), the copyright and terms of service of that source should be provided.
        \item If assets are released, the license, copyright information, and terms of use in the package should be provided. For popular datasets, \url{paperswithcode.com/datasets} has curated licenses for some datasets. Their licensing guide can help determine the license of a dataset.
        \item For existing datasets that are re-packaged, both the original license and the license of the derived asset (if it has changed) should be provided.
        \item If this information is not available online, the authors are encouraged to reach out to the asset's creators.
    \end{itemize}

\item {\bf New assets}
    \item[] Question: Are new assets introduced in the paper well documented and is the documentation provided alongside the assets?
    \item[] Answer: \answerYes{} 
    \item[] Justification: The paper clearly describes our proposed algorithm in \cref{sec:algorithm}. The codes of our proposed algorithm are provided in a ZIP file.
    \item[] Guidelines:
    \begin{itemize}
        \item The answer NA means that the paper does not release new assets.
        \item Researchers should communicate the details of the dataset/code/model as part of their submissions via structured templates. This includes details about training, license, limitations, etc. 
        \item The paper should discuss whether and how consent was obtained from people whose asset is used.
        \item At submission time, remember to anonymize your assets (if applicable). You can either create an anonymized URL or include an anonymized zip file.
    \end{itemize}

\item {\bf Crowdsourcing and research with human subjects}
    \item[] Question: For crowdsourcing experiments and research with human subjects, does the paper include the full text of instructions given to participants and screenshots, if applicable, as well as details about compensation (if any)? 
    \item[] Answer: \answerNA{} 
    \item[] Justification: Our paper does not involve crowdsourcing nor research with human subjects.
    \item[] Guidelines:
    \begin{itemize}
        \item The answer NA means that the paper does not involve crowdsourcing nor research with human subjects.
        \item Including this information in the supplemental material is fine, but if the main contribution of the paper involves human subjects, then as much detail as possible should be included in the main paper. 
        \item According to the NeurIPS Code of Ethics, workers involved in data collection, curation, or other labor should be paid at least the minimum wage in the country of the data collector. 
    \end{itemize}

\item {\bf Institutional review board (IRB) approvals or equivalent for research with human subjects}
    \item[] Question: Does the paper describe potential risks incurred by study participants, whether such risks were disclosed to the subjects, and whether Institutional Review Board (IRB) approvals (or an equivalent approval/review based on the requirements of your country or institution) were obtained?
    \item[] Answer: \answerNA{} 
    \item[] Justification: Our paper does not involve crowdsourcing nor research with human subjects.
    \item[] Guidelines:
    \begin{itemize}
        \item The answer NA means that the paper does not involve crowdsourcing nor research with human subjects.
        \item Depending on the country in which research is conducted, IRB approval (or equivalent) may be required for any human subjects research. If you obtained IRB approval, you should clearly state this in the paper. 
        \item We recognize that the procedures for this may vary significantly between institutions and locations, and we expect authors to adhere to the NeurIPS Code of Ethics and the guidelines for their institution. 
        \item For initial submissions, do not include any information that would break anonymity (if applicable), such as the institution conducting the review.
    \end{itemize}

\item {\bf Declaration of LLM usage}
    \item[] Question: Does the paper describe the usage of LLMs if it is an important, original, or non-standard component of the core methods in this research? Note that if the LLM is used only for writing, editing, or formatting purposes and does not impact the core methodology, scientific rigorousness, or originality of the research, declaration is not required.
    \item[] Answer: \answerNA{} 
    \item[] Justification: Our core method development in this research does not involve any usage of LLMs.
    \item[] Guidelines:
    \begin{itemize}
        \item The answer NA means that the core method development in this research does not involve LLMs as any important, original, or non-standard components.
        \item Please refer to our LLM policy (\url{https://neurips.cc/Conferences/2025/LLM}) for what should or should not be described.
    \end{itemize}

\end{enumerate}

\newpage
\appendix

\section{Proof of \cref{thm:ptow}}\label{app:A}
In this section we present the proof for the necessary properties of effective Pareto optimal arms. Before proving this theorem, we use the following definition that denotes the convex hull of arbitrary set. 
\begin{definition}
    Let $\Mcal\subset\RR^L$ be a set of mean reward vector $\mub_a$. For all $a\in\Acal$. Define the convex hull of a set $\Mcal$ as $\normalfont\textbf{Conv}(\Mcal)$. 
\end{definition}

By definition, for any $\beta = (\beta_a)_{a\in\Acal}\in\Scal^{|\Acal|}$, we have 
\begin{equation*}
     \sum_{a\in\Acal} \beta_a\mub_a \in \normalfont\textbf{Conv}(\Mcal)    
\end{equation*}

This convex hull covers all the convex combination of mean reward vectors of every arms. 
We note that, in the convex set $\normalfont\textbf{Conv}(\Mcal)$, the mean reward vector of the effective Pareto optimal arm satisfies the Pareto optimality in $\normalfont\textbf{Conv}(\Mcal)$. In other words, for all $a_*\in \Ccal^*$, we have
\begin{equation}\label{eq:epo}
    \mub_{a_*}\not\prec \mub, \qquad \forall \mub\in\normalfont\textbf{Conv}(\Mcal)\setminus\{\mub_{a_*}\}
\end{equation}
However, for the arm $a\in\Acal\setminus\Ccal^*$, there exists $\mub\in\normalfont\textbf{Conv}(\Mcal)$ satisfying
\begin{equation*}
    \mub_a\prec\mub
\end{equation*}
The arm that is effective Pareto optimal, satisfies the Pareto optimality in $\normalfont\textbf{Conv}(\Mcal)$ (see \cref{def:CPO}).

Other than the Pareto optimality, the next definition describes the optimality of one specific objective, with constraint on the other objectives.

\begin{definition}[$\epsilonb$-constraint optimal]
    Let $\epsilonb_\ell$ be $L-1$ dimensional arbitrary constraint vector
    \begin{equation*}
        \boldsymbol{\epsilon}_\ell = \begin{bmatrix}
        \epsilon\obj{1} & ... & \epsilon\obj{\ell-1} & \epsilon\obj{\ell+1} & ... & \epsilon\obj{L}
        \end{bmatrix}^\top \in\RR^{L-1}
    \end{equation*}
    Given the constraint vector $\epsilonb_\ell$, the $\boldsymbol{\epsilon}_\ell$-constraint optimal vector among the set $\normalfont\textbf{Conv}(\Mcal)$, denoted $\mub_{\ell_*}$, is defined by
    \begin{equation*}
        \mub_{\ell_*} = \argmax_{\mub \in \normalfont\textbf{Conv}(\Mcal)}\{\mu\obj{l}\mid\mu\obj{k}\ge\epsilon\obj{k}\text{ for all } k = 1,2,...,L, k\ne \ell\} 
    \end{equation*}
    The vector is $\boldsymbol\epsilonb$-constraint optimal (denoted $\mub_*$) if, for all $\ell\in[L]$, there exists constraint vector $\epsilonb_{\ell}$ such that the vector $\mub_*$ is $\boldsymbol{\epsilon}_\ell$-constraint optimal vector.
\end{definition}
The constraint vector $\boldsymbol\epsilonb_\ell$ is the lower bound values, such that the vector $\mub$ dominates the constraint vector, except objective $\ell$.
Among those vector $\mub$ satisfying constraint, the $\boldsymbol{\epsilon}_\ell$-constraint optimal vector is the one that has maximum value in objective $\ell$.
The $\boldsymbol\epsilonb$-constraint optimal vector is such constraint vector $\boldsymbol\epsilonb_\ell$ exists, as to be $\boldsymbol\epsilonb_\ell$ constraint optimal, for all $\ell\in[L]$.

The next lemma shows the equivalence between $\epsilonb$-constraint optimality and Pareto optimality. 
\begin{lemma}\label{lem:ptoc}
    The vector $\mub_*\in\normalfont\textbf{Conv}(\Mcal)$ is Pareto optimal if and only if it is $\epsilonb$-constraint optimal.
\end{lemma}
\begin{proof}
    ($\Longrightarrow$) Let $\mub_*$ be Pareto optimal. Assume it is not $\boldsymbol{\epsilon}_\ell$-constraint optimal for some $\ell$. Let the constraint vector be $\epsilon\obj{k} = \mu_{*}\obj{k}$ for $k=1,...,L,k\ne \ell$. Since it is not $\boldsymbol{\epsilon}_\ell$-constraint optimal, then there exists vector $\dot\mub$ such that
    $\mu_{*}\obj{k} \le \dot\mu\obj{k}$ for $k=1,...,L$ and $\mu_{*}\obj{\ell} < \dot\mu\obj{\ell}$. Since $\dot\mub$ exists and dominates $\mub_*$, this contradicts the definition of Pareto optimality.
    
    ($\Longleftarrow$) Let $\mub_*$ be $\epsilonb$-constraint optimal. Suppose the constraint vector is defined as $\epsilon\obj{\ell} = \mu_*\obj{\ell}$ for all $\ell\in[L]$. Since $\mub_*$ is $\boldsymbol{\epsilon}_\ell$-constraint optimal for every $\ell = 1,...,L$, there is no other $\mub\in\normalfont\textbf{Conv}(\Mcal)$ satisfying $\mub_{*}\obj{\ell} < \dot\mub\obj{\ell}$ and $\mub_{*}\obj{k} \le \dot\mub_{}\obj{k}$ when $k\ne \ell$, for every $\ell = 1,...,L$. This holds the definition of Pareto optimality.
\end{proof}
Above lemma demonstrates that every Pareto optimal arm is also $\epsilonb$-constraint optimal. This equivalence also holds for non-convex set. But in the convex set $\normalfont\textbf{Conv}(\Mcal)$, the mean reward vector effective Pareto optimal arm satisfies the Pareto optimality in $\normalfont\textbf{Conv}(\Mcal)$, hence, it also satisfies $\epsilonb$-constraint optimal. It is enough to show that, for any $\epsilonb$-constraint optimal vector $\mub_*\in\normalfont\textbf{Conv}(\Mcal)$, there exists weight vector $\wb\in\Scal^{L}$ satisfying
\begin{equation*}
    \mub_* = \argmax_{\mub\in\normalfont\textbf{Conv}(\Mcal)}\wb^\top\mub
\end{equation*}

To prove above equation, we prove some of the lemmas that are useful for our proof.

\begin{lemma}\label{lem:a1}
    let $\Omega$ be non-empty convex set in $\RR^L$, not containing origin. Then there exists a vector $\wb\in\Scal^L$ such that $\wb^\top\mub \ge 0$ holds for all $\mub\in\Omega$.
\end{lemma}
\begin{proof}
    for $\mub_1, \mub_2, ..., \mub_m\in\Omega$, define matrix and arbitrary vector as
    \begin{equation*}
        M =\begin{bmatrix}
            \mub_1 & \mub_2 & ... & \mub_m
        \end{bmatrix}^\top\in\RR^{m\times L}, \quad \beta\in \Scal^m.
    \end{equation*}
    by convexity of set $\Omega$, we have $M^\top\beta \in \Omega$, but $0 \not\in\Omega$. So, there is no solution $\beta$, satisfying
    \begin{equation*}
        M^\top\beta =  0,\quad \beta\in \Scal^m.
    \end{equation*}
    The solution still do not exist even if we remove the constraint $\|\beta\|_1 = 1$. By \cref{lem:Gordan}, the second condition of Gordan's theorem does not hold. Hence, there exist $L$-dimensional vector $\wb$ that $\wb^\top\mub_i > 0$ holds for all $i = 1,...,m$. Since $\wb$ is non-zero vector, we can take $\wb$ as $\sum_{\ell\in[L]}|w\obj{\ell}| = 1$. Define the set
    \begin{equation*}
        V_{\mub_i} = \{\wb\in\RR^L\mid \sum_{\ell\in[L]}|w\obj{\ell}| = 1, \wb^\top\mub_i \ge 0 \}.
    \end{equation*}
    Then we can write
    \begin{equation*}
        \bigcap_{i=1,...,m} V_{\mub_i} \not= \emptyset.
    \end{equation*}
    Each set $V_{\mub_i}$ is closed and bounded, hence, it is compact set. Since $\mub_i$ was arbitrary chosen, the collection $(V_{\mub})_{\mub}$ satisfies finite intersection property. So, we have
    \begin{equation*}
        \bigcap_{\mub\in\Omega} V_{\mub} \not= \emptyset.
    \end{equation*}
\end{proof}
\begin{lemma} \label{lem:a2}
    Let $\Omega$ be non-empty convex set in $\RR^L$, such that the vector $\mub\in\Omega$ with all negative entries do not exist. Then, there exist vector $\wb\in S^L$ such that $\wb^\top\mub \ge0$ holds for all $\mub \in \Omega$.
\end{lemma}
\begin{proof}
    For a vector $\mub \in \Omega$, define the set 
    \begin{align*}
        \Bcal_{\mub} &=\{\yb\in\RR^L|y\obj{\ell} > \mu\obj{\ell},\forall \ell\in[L]\},    \\
        \Bcal &= \bigcup_{\mu\in\Omega}\Bcal_\mu.
    \end{align*}
    If origin is in $\Bcal$, then there exists $\mub$ that $0 > \mu\obj{\ell}$ holds for all $\ell\in[L]$, which contradicts the assumption. If $\yb_1\in\Bcal_{\mub_1}$, $\yb_2\in\Bcal_{\mub_2}$, we have 
    \begin{equation*}
        \gamma\yb_1+(1-\gamma)\yb_2 \in \Bcal_{\gamma\mub_1 + (1-\gamma)\mub_2} \subset \Bcal
    \end{equation*}
    for $\gamma\in [0,1]$. Hence, $\Bcal$ is convex set. By \cref{lem:a1}, there exist vector $\wb$, satisfying $\wb^\top \yb \ge 0$ for all $\yb \in \Bcal$. If the vector has negative entry  $w\obj{\ell} < 0$, we can choose $\yb\in\Bcal$ with large $y\obj{\ell}$ so that $\wb^\top \yb < 0$. Hence, we must have $w\obj{\ell} \ge 0$ for all $\ell\in[L]$. Also, since $\wb$ is non-zero vector, we can restrict the vector in unit $L$-dimensional simplex.
    We now prove $\wb^\top \mub \ge 0$ for all $\mub \in \Omega$. For any positive real $\epsilon > 0$, we have $\mub + \epsilon\one\in\Bcal$. If there exist $\delta > 0,\mub$ with $\wb^\top\mub = -\delta$, we can choose $\epsilon < \delta$, so that
    \begin{equation*}
        \wb^\top(\mub + \epsilon\one) = -\delta + \epsilon < 0.
    \end{equation*}
    Hence, we must have vector $\wb$ that satisfies $\wb^\top \mub \ge 0$ for all $\mub \in \Omega$, and $\wb \in \Scal^L$
\end{proof}

The next lemma is the revision of Generalized Gordan's Theorem. 

\begin{lemma}[Generalized Gordan's Theorem]\label{lem:genGordan}
    Let $\Omega$ be non-empty convex set. Either one of the following statements holds, but not both.
    \begin{enumerate}
        \item There exists $\mub\in\Omega$ whose entries are all negative.
        \item There exists $L$-dimensional vector $\wb\in \Scal^L$ satisfying $\wb^\top\mub \ge 0$ for all $\mub \in \Omega$.
    \end{enumerate}
\end{lemma}
\begin{proof}
    ($\bar1\Longrightarrow 2$) proof follows by \cref{lem:a2}. \\
    ($2\Longrightarrow \bar1$) If $\mub$ is negative vector, we must have $\wb^\top\mub < 0$ for all $\wb \in \Scal^L$.
\end{proof}

\begin{lemma}\label{lem:conv}
    For any $\wb\in \Scal^L$, let $\Acal^*_{\wb}$ be set of optimal arm, that has optimal weight sum reward given weight vector $\wb$, i.e.,
    \begin{equation*}
        \Acal^*_{\wb} = \arg\max_{a\in\Acal} \wb^\top\mub_a.
    \end{equation*}
    Then there exists effective Pareto optimal arm $a_*$, such that $a_*\in \Acal^*_w$. 
\end{lemma}

\begin{proof}
    Suppose there exist $\wb$ that $a_*\not\in\Acal^*_{\wb}$ for any $a_*\in\Ccal^*$. Let $\bar a\in\Acal^*_{\wb}/\Ccal^*$ that maximizes $\wb^\top\mub_a$. Since $\bar a\not\in\Ccal^*$, for every effective Pareto optimal arm $a_*\in\Ccal^*$, there exist $\beta_{a_*}\ge0$ satisfying
    \begin{equation*}
        \mub_{\bar a} \prec \sum_{a_*\in\Ccal^*}\beta_{a_*}\mub_{a_*}, \quad \sum_{a_*\in\Ccal^*}\beta_{a_*}=1.
    \end{equation*}
    Since $\wb$ is vector with non-negative entries, we have
    \begin{equation*}
        \wb^\top\mub_{\bar a} =\sum_{a_*\in\Ccal^*}\beta_{a_*}\wb^\top\mub_{\bar a} \le \sum_{a_*\in\Ccal^*}\beta_{a_*}\wb^\top \mub_{a_*}.
    \end{equation*}
    We have, at least, one Pareto optimal arm with 
    \begin{equation*}
        \wb^\top\mub_{\bar a} \le \wb^\top \mub_{a_*}
    \end{equation*}
    Such existence of $a_*$ is guaranteed with existence of $\beta_{a_*}>0$. 
\end{proof}
Now, we begin the proof of \pref{thm:ptow}

\begin{proof}
    Suppose $a_*$ is effective Pareto optimal. By \cref{eq:epo}, for any $\mub\in\textbf{Conv}(\Mcal)\setminus\{\mub_{a_*}\}$, we have $\mub_{a_*} \not\prec \mub$, hence, the vector $\mub_{a_*}$ satisfies the Pareto optimality in $\textbf{Conv}(\Mcal)$. By \cref{lem:ptoc}, $\mub_{a_*}$ is also $\epsilonb$-constraint optimal, with $\epsilon\obj{\ell} = \mu_{a_*}\obj{\ell}$, such that no vector $\mub\in\textbf{Conv}(\Mcal)\setminus\{\mub_{a_*}\}$ satisfies $\mu\obj{k}_{a_*}-\mu\obj{k}_{h} \le 0$ for $k = 1,...,L$ and $\mu\obj{\ell}_{a_*}-\mu\obj{\ell}_{h} < 0$ for some $\ell$. By \cref{lem:genGordan}, since the set $\textbf{Conv}(\Mcal)$ is convex, the first statement does not hold. There exists $L$-dimensional vector $\wb\in \Scal^L$ satisfying $\wb^\top(\mub_{a_*}-\mub_{}) \ge 0$ for all $\mub \in \textbf{Conv}(\Mcal)$. Hence, we have
    \begin{equation*}
        a_* = \argmax_{a\in\Acal} \wb^\top\mub_{a}
    \end{equation*}

    Conversly, Suppose $a^*_{\wb} = \arg\max_{a\in\Acal} \wb^\top\mub_a$ is unique arm. By \cref{lem:conv}, the existence of effective Pareto optimal arm implies $a^*_{\wb}\in\Ccal^*$.
\end{proof}
\section{Analysis of \texttt{MOLB-TS}}\label{app:B}
In this section, we provide the analysis of the worst-case regret of algorithm \texttt{MOLB-TS}.

We begin with the proof of \cref{lem:opt}.
\subsection{Proof of \cref{lem:opt}}\label{prf:opt}
\begin{proof}
    The parameters $(\tilde\theta_{t,m}\obj{\ell})_{m\in[M]}$ are sampled from Gaussian distribution $\Ncal(\hat\theta_t\obj{\ell}, c_{1,t}^2 V_t^{-1})$. Then for any given $d$-dimensional vector $x_{t,a}$, we can rewrite this probability distribution as
    \begin{equation*}
        x_{t,a}^\top\tilde\theta_{t,m}\obj{\ell}\sim\Ncal(x_{t,a}^\top\hat\theta_t\obj{\ell}, c_{1,t}^2 \|x_{t,a}\|_{V_t^{-1}}^2 ).
    \end{equation*}
    Also, we can see the probability of the event $\dot\Ecal_{t,a}\obj{\ell}$ as
    \begin{align*}
        \PP(\dot\Ecal_{t,a}\obj{\ell}) &= \PP\{ \exists m\in[M]: x_{t,a}^\top(\tilde\theta_{t,m}\obj{\ell}-\hat\theta_{t}\obj{\ell})\ge c_{1,t}\|x_{t,a}\|_{V_t^{-1}}\} \\
        &= \PP\{\exists m\in[M]:\eta_{m} \ge 1\}
    \end{align*}
    where $(\eta_{m})_{m\in[M]}$ is sampled from standard normal distribution $\Ncal(0,1)$. For $M=1$, the probability of optimism $\PP(\dot\Ecal_{t,a}\obj{\ell})$ is bounded below by $p$. The probability of optimism, satisfying at least one sampled parameter, is bounded by
    \begin{equation*}
        \PP(\dot\Ecal_{t,a}\obj{\ell}) \ge 1-(1-p)^M.
    \end{equation*}
    These optimism events between different objectives are independent. Let the event $\dot\Ecal_{t,a}$ be defined as
    \begin{equation*}
        \dot\Ecal_{t,a} = \bigcap_{\ell\in[L]}\dot\Ecal_{t,a}\obj{\ell}
    \end{equation*}
    The event $\dot\Ecal_{t,a}$ is the optimism satisfying for all objectives $\ell\in[L]$. Hence, we have
    \begin{equation*}
        \PP(\dot\Ecal_{t,a}) \ge (1-(1-p)^M)^L.
    \end{equation*}
    To have $\PP(\dot\Ecal_{t,a}) \ge p$, we need to choose large enough $M$ so that
    \begin{align*}
        (1-(1-p)^M)^L &\ge p.         
    \end{align*}
    Rearrange the equation, we get
    \begin{align*}
        M&\ge\frac{\log(1-p^{1/L})}{\log(1-p)} \\
        &= \frac{\log(1-p)-\log(1+p^{1/L}+p^{2/L}... +p^{(L-1)/L})}{\log(1-p)} \\
        &= 1 + \frac{\log(1+p^{1/L}+p^{2/L}... +p^{(L-1)/L})}{\log \frac{1}{1-p}} 
    \end{align*}
    With sampling by Gaussian distribution, we have $p=0.15$. However, the probability $p$ can be different by choosing different sampling distribution. But, as long as $p\in[0,1)$, we have
    \begin{equation*}
        1 + \frac{\log L}{\log \frac{1}{1-p}}\ge
        1 + \frac{\log(1+p^{1/L}+p^{2/L}... +p^{(L-1)/L})}{\log \frac{1}{1-p}}
    \end{equation*}
    Hence, by choosing $M$ as
    \begin{equation*}
        M \ge \frac{\log L}{\log \frac{1}{1-p}},
    \end{equation*}
    we get $\PP(\dot\Ecal_{t,a}) \ge p$
\end{proof}

\cref{lem:opt} shows the minimum number of $M$ for the inequality $\PP(\dot\Ecal_{t,a}) \ge p$ to hold. 

\subsection{Proof of \cref{thm:regret}}

\textbf{Dependence on $\log L$.} Before proving \cref{thm:regret}, we define the events $\hat\Ecal_t, \tilde\Ecal_t$ such that the true parameters $\theta_*\obj{\ell}$ and all sampled parameters $(\tilde\theta_{t,m}\obj{\ell})_{m\in[M]}$ are close enough to the RLS estimate parameters $\hat\theta_t\obj{\ell}$ for all objectives, respectively.
\begin{align*}
    \hat\Ecal_t &:=\{\forall \ell \in[L] : \|\theta_*\obj{\ell}-\hat\theta_t\obj{\ell}\|_{V_t}\le c_{1,t}(\delta)\}, \\
    \tilde\Ecal_t &:= \{ \forall m\in[M],\forall \ell\in[L]: \|\tilde\theta_{t,m}\obj{\ell}-\hat\theta_t\obj{\ell}\|_{V_t}\le c_{2,t}(\delta)\},
\end{align*}
where $c_{1,t}(\delta)$ and $ c_{2,t}(\delta)$ are defined as
\begin{equation*}
    \begin{aligned}
        c_{1,t}(\delta) &:= R\sqrt{d\log\left(\frac{1+(t-1)/(\lambda d)}{\delta/L}\right)}+\lambda^{1/2},\\
        c_{2,t}(\delta) &:= c_{1,t}(\delta)\sqrt{2d\log\frac{2LMdT}{\delta}}.
    \end{aligned}
\end{equation*}    
Let $\hat\Ecal = \bigcap_{t\ge0}\hat\Ecal_t$. By \cref{lem:abbasi1}, we have $\PP(\hat\Ecal) \ge 1-\delta$, and by \cref{lem:abeille1}, on event $\hat\Ecal $, we have $\PP_t(\tilde\Ecal_t) = \PP(\tilde\Ecal_t\mid\Fcal_t)\ge 1-\nicefrac{\delta}{T}$. The high probability bounds $c_{1,t}(\delta)$ and $c_{2,t}(\delta)$ increased by a factor of $\log L$ due to the union bound over the number of objectives. Consequently, the regret inevitably depends on $\log L$ as this factor arises from the concentration bounds required to hold uniformly over all objectives.

\textbf{Bounding the sub-optimality gap.} We now prove the following lemma, which bounds the conditional expectation of the regret of \texttt{MOLB-TS} at round $t$ given the historical information up to that point.
\begin{lemma}\label{lem:bound}
    For any filtration $\Fcal_{t-1}$, on event $\hat\Ecal$, we have
    \begin{equation*}
        \EE_t[\Delta_{a_t}^{EPR}]\le\left(1+\frac{2}{0.15-\frac{\delta}{T}}\right)\left(c_{1,t}(\delta)+c_{2,t}(\delta)\right)\EE_t[\|x_{t, a_t}\|_{V_t^{-1}}] + \frac{\delta}{T}\Delta_{max}
    \end{equation*}
\end{lemma}
\begin{proof}
Let $(\beta_{t,a_*})_{a_*\in\Ccal^*_t}$ be one that maximizes $\Delta_{a_t}^{EPR}$.
\begin{align*}
    \Delta_{a_t}^{EPR} &= \max_{\beta\in\Scal^{|\Ccal^*_t|}}\min_{\ell\in[L]}\Biggl\{\left(\sum_{a_*\in\mathcal C^*_t} \beta_{a_*} x_{t,a_*}^\top \theta_*^{(\ell)}\right)-x_{t,a_t}^\top\theta_*^{(\ell)}\Biggr\} \\
    &=\min_{\ell\in[L]}\Biggl\{\left(\sum_{a_*\in\mathcal C^*_t} \beta_{t,a_*} x_{t,a_*}^\top \theta_*^{(\ell)}\right)-x_{t,a_t}^\top\theta_*^{(\ell)}\Biggr\} \\
\end{align*}

Let $\wb_t$ be the weight vector in unit $L$-simplex sampled, described in \cref{sec:bound}. Then, 
\begin{align*}
    \Delta_{a_t}^{EPR} &=\min_{\ell\in[L]}\Biggl\{\left(\sum_{a_*\in\mathcal C^*_t} \beta_{t,a_*} x_{t,a_*}^\top \theta_*^{(\ell)}\right)-x_{t,a_t}^\top\theta_*^{(\ell)}\Biggr\} \\
    &\le \sum_{\ell\in[L]}w^{(\ell)}_t\left(\left(\sum_{a_*\in\mathcal C^*_t} \beta_{t,a_*} x_{t,a_*}^\top \theta_*^{(\ell)}\right)-x_{t,a_t}^\top\theta_*^{(\ell)} \right)\\
    &=\sum_{a_*\in\Ccal^*_t} \beta_{t,a_*} x_{t,a_*}^\top \left(\sum_{\ell\in[L]}w^{(\ell)}_t\theta_*^{(\ell)}\right)-x_{t,a_t}^\top\left(\sum_{\ell\in[L]}w^{(\ell)}_t\theta_*^{(\ell)} \right).
\end{align*}
By \cref{thm:ptow}, there exists $\bar a_* \in \Ccal^*$ satisfying 
\begin{equation*}
    \bar a_*=\argmax_{a\in\mathcal A}x_{t,a}^\top\sum_{\ell\in[L]}w^{(\ell)}_t\theta_*^{(\ell)}.
\end{equation*}
Hence, we have
\begin{align}
    \Delta_{a_t}^{EPR} &\le \sum_{a_*\in\Ccal^*_t} \beta_{t,a_*} x_{t,a_*}^\top \left(\sum_{\ell\in[L]}w^{(\ell)}_t\theta_*^{(\ell)}\right)-x_{t,a_t}^\top\left(\sum_{\ell\in[L]}w^{(\ell)}_t\theta_*^{(\ell)} \right) \\
    &\le x_{t, \bar a_*}^\top \left(\sum_{\ell\in[L]}w^{(\ell)}_t\theta_*^{(\ell)}\right)-x_{t,a_t}^\top\left(\sum_{l\in[L]}w^{(\ell)}_t\theta_*^{(\ell)} \right). 
    \label{eq:1}
\end{align}
From $M$ multiple sampled parameters, we define $\tilde\theta_{a,t}\obj{\ell}$ as optimal sampled parameter with arm $a$, i.e.,
\begin{equation*}
        \tilde\theta_{a,t}\obj{\ell} = \argmax_{\tilde\theta_{t,m}}\{x_{t,a}^\top\tilde\theta_{t,1}\obj{\ell}, x_{t,a}^\top\tilde\theta_{t,2}\obj{\ell}, ..., x_{t,a}^\top\tilde\theta_{t,M}\obj{\ell}\}.
\end{equation*}
At round $t$, the arm $a$ is evaluated with the sampled parameters $\tilde\theta_{a,t}\obj{\ell}$ for all $\ell\in[L]$.
As we described in \cref{sec:bound}, we can write $\bar a_*, a_t$ as
    \begin{align*}
        \bar a_* &=\argmax_{a\in\mathcal A}x_{t,a}^\top\sum_{l\in[L]}w^{(\ell)}_t\theta_*^{(\ell)}, \\
        a_t &= \argmax_{a\in\mathcal A}x_{t,a}^\top\sum_{l\in[L]}w^{(\ell)}_t\tilde\theta_{a,t}^{(\ell)}.
    \end{align*}
 Let $c_t(\delta)=c_{1,t}(\delta)+c_{2,t}(\delta)$. We separated arms into two sets with given weight vector, saturated and unsaturated ~\cite{agrawal2013thompson}. 
    \begin{itemize}
        \item $\Bcal_t$ : set of saturated arms, that is, for all $a\in\Bcal_t$, we have 
        \begin{align*}
            x_{t, \bar a_*}^\top \left(\sum_{\ell\in[L]}w^{(\ell)}_t\theta_*^{(\ell)}\right)-x_{t,a}^\top\left(\sum_{l\in[L]}w^{(\ell)}_t\theta_*^{(\ell)} \right)  > c_t(\delta)\| x_{t,a}\|_{V_t^{-1}}.
        \end{align*}
        \item $\overline\Bcal_t$ : set of unsaturated arms, that is, for all $a\in\overline\Bcal_t$, we have 
        \begin{align*}
            x_{t, \bar a_*}^\top \left(\sum_{\ell\in[L]}w^{(\ell)}_t\theta_*^{(\ell)}\right)-x_{t,a}^\top\left(\sum_{l\in[L]}w^{(\ell)}_t\theta_*^{(\ell)} \right)  \le c_t(\delta)\| x_{t,a}\|_{V_t^{-1}}.
        \end{align*}
    \end{itemize}
    Note that $\wb_t$ is random variable, since the arm $a_t$ is uniform randomly selected from $\tilde\Ccal_t$. 
    Hence, those sets of saturated and unsaturated arms ($\Bcal_t, \bar\Bcal_t$) are not fixed.
    Let $\bar a_t = \argmin_{a\in\bar\Bcal_t}\|x_{t,a}\|_{V_t^{-1}}$ be arm in $\bar\Bcal_t$ with smallest matrix norm.
From \cref{eq:1}, bounding the sub-optimality gap $\Delta_{a_t}^{EPR}$ on event $\hat\Ecal$ and $\tilde\Ecal_t$, we have
    \begin{align*}
        \Delta_{a_t}^{EPR} &\le x_{t, \bar a_*}^\top\left(\sum_{\ell\in[L]}w^{(\ell)}_t\theta^{(\ell)}_*\right)-x_{t,a_t}^\top\left(\sum_{\ell\in[L]}w^{(\ell)}_t\theta_*^{(\ell)}\right) \\
        &= x_{t,\bar a_*}^\top\left(\sum_{\ell\in[L]}w^{(\ell)}_t\theta^{(\ell)}_*\right)+x_{t,\bar a_t}^\top\left(\sum_{\ell\in[L]}w^{(\ell)}_t\theta^{(\ell)}_*\right) \\
        &\quad - x_{t, \bar a_t}^\top\left(\sum_{\ell\in[L]}w^{(\ell)}_t\theta^{(\ell)}_*\right)-x_{t, a_t}^\top\left(\sum_{\ell\in[L]}w^{(\ell)}_t\theta_*^{(\ell)}\right) 
    \end{align*}
For any arm $a$, by the Cauchy-Schwarz inequality of matrix norm, we have
    \begin{align*}
        \left|x_{t,a}^\top\left(\sum_{\ell\in[L]}w^{(\ell)}_t\tilde\theta^{(\ell)}_{a,t}\right) - x_{t, a}^\top\left(\sum_{\ell\in[L]}w^{(\ell)}_t\theta^{(\ell)}_*\right)\right|
        &\le \|x_{t,a}\|_{V_t^{-1}}\left(\sum_{\ell\in[L]}w_t\obj{\ell}\|\tilde\theta^{(\ell)}_{a,t}-\theta^{(\ell)}_*\|_{V_t}\right).
    \end{align*}
    And by triangle inequality of norm, we have
    \begin{align*}
        \sum_{\ell\in[L]}w_t\obj{\ell}\|\tilde\theta^{(\ell)}_{a,t}-\theta^{(\ell)}_*\|_{V_t}
         &\le \sum_{\ell\in[L]}w_t\obj{\ell}\left(\|\tilde\theta^{(\ell)}_{a,t}-\hat\theta\obj{\ell}_t\|_{V_t} + \|\hat\theta\obj{\ell}_t- \theta^{(\ell)}_*\|_{V_t}\right).\\
    \end{align*}
    And lastly, on event $\hat\Ecal$ and $\tilde\Ecal_t$, we get
    \begin{align*}
        \|\tilde\theta^{(\ell)}_{a,t}-\hat\theta\obj{\ell}_t\|_{V_t} + \|\hat\theta\obj{\ell}_t- \theta^{(\ell)}_*\|_{V_t} \le (c_{1,t}(\delta)+c_{2,t}(\delta)) = c_{t}(\delta)
    \end{align*}
    In total, we get
    \begin{align*}
        \left|x_{t,a}^\top\left(\sum_{\ell\in[L]}w^{(\ell)}_t\tilde\theta^{(\ell)}_{a,t}\right) - x_{t, a}^\top\left(\sum_{\ell\in[L]}w^{(\ell)}_t\theta^{(\ell)}_*\right)\right| \le \left(\sum_{\ell\in[L]}w_t\obj{\ell}\right)c_{t}(\delta)\|x_{t,a}\|_{V_t^{-1}} .
    \end{align*}
    Hence, on event $\hat\Ecal$ and $\tilde\Ecal_t$, 
    \begin{align*}
         \Delta_{a_t}^{EPR}&\le x_{t, \bar a_*}^\top\left(\sum_{\ell\in[L]}w^{(\ell)}_t\theta^{(\ell)}_*\right) 
         +x_{t, \bar a_t}^\top\left(\sum_{\ell\in[L]}w^{(\ell)}_t\tilde\theta^{(\ell)}_{\bar a_t,t}\right)
         +\left(\sum_{\ell\in[L]}w^{(\ell)}_t\right)c_t(\delta)\|x_{t, \bar a_t}\|_{V_t^{-1}}\\
         &\quad -x_{t,\bar a_t}^\top\left(\sum_{\ell\in[L]}w^{(\ell)}_t\theta^{(\ell)}_*\right)
         -x_{t,a_t}^\top\left(\sum_{\ell\in[L]}w^{(\ell)}_t\tilde\theta^{(\ell)}_{a_t,t}\right)
         + \left(\sum_{\ell\in[L]}w^{(\ell)}_t\right)c_t(\delta)\|x_{t, a_t}\|_{V_t^{-1}} \\
        &= x_{t,\bar a_*}^\top\left(\sum_{\ell\in[L]}w^{(\ell)}_t\theta^{(\ell)}_*\right)
        +x_{t,\bar a_t}^\top\left(\sum_{l\in[L]}w^{(\ell)}_t\tilde\theta^{(\ell)}_{\bar a_t,t}\right)
        +c_t(\delta)\|x_{t,\bar a_t}\|_{V_t^{-1}}\\
        &\quad -x_{t,\bar a_t}^\top\left(\sum_{\ell\in[L]}w^{(\ell)}_t\theta^{(\ell)}_*\right) 
        -x_{t,a_t}^\top\left(\sum_{\ell\in[L]}w^{(\ell)}_t\tilde\theta^{(\ell)}_{a_t,t}\right)        
        + c_t(\delta)\|x_{t, a_t}\|_{V_t^{-1}}.
    \end{align*}
    Since
    \begin{align*}
        a_t = \argmax_{a\in\mathcal A}x_{t,a}^\top\sum_{l\in[L]}w^{(\ell)}_t\tilde\theta_{a,t}^{(\ell)},
    \end{align*}
    we have
    \begin{align*}
        \Delta_{a_t}^{EPR} &\le x_{t,\bar a_*}^\top\left(\sum_{\ell\in[L]}w^{(\ell)}_t\theta^{(\ell)}_*\right)-x_{t,\bar a_t}^\top\left(\sum_{\ell\in[L]}w^{(\ell)}_t\theta^{(\ell)}_*\right)  +c_t(\delta)\|x_{t, \bar a_t}\|_{V_t^{-1}}+ c_t(\delta)\|x_{t, a_t}\|_{V_t^{-1}} \\
        &\le 2c_t(\delta)\|x_{t,\bar a_t}\|_{V_t^{-1}}+ c_t(\delta)\|x_{t, a_t}\|_{V_t^{-1}}, \\
    \end{align*}
    where the last inequality holds since $\bar a_t$ is unsaturated arm. 
    This inequality holds on event $\hat\Ecal$ and $\tilde\Ecal_t$. 
    Define the conditional probability $\PP_t = \PP(\cdot \mid\Fcal_t)$.
    Then, on event $\hat\Ecal$, we have
    \begin{align*}
        \EE_t[\Delta_{a_t}^{CPR}] &\le\mathbb E_t[2c_t(\delta)\|x_{t,\bar a_t}\|_{V_t^{-1}} + c_t(\delta)\|x_{t, a_t}\|_{V_t^{-1}}] + \mathbb (1-\PP_t\{\tilde\Ecal_t\})\Delta_{max} \\
        &\le\mathbb E_t[2c_t(\delta)\|x_{t,\bar a_t}\|_{V_t^{-1}} + c_t(\delta)\|x_{t,a_t}\|_{V_t^{-1}}] + \frac{\delta}{T}\Delta_{max}
    \end{align*}
    We bound the term $\EE_t[\|x_{t,\bar a_t}\|_{V_t^{-1}}]$ with $\|x_{t, a_t}\|_{V_t^{-1}}$. We have
    \begin{align*}
        \EE_t[\|x_{t,a_t}\|_{V_t^{-1}}] &\ge\ \EE_t[\|x_{t, a_t}\|_{V_t^{-1}}\mid a_t\in \overline \Bcal_t]\mathbb P_t\{a_t\in \overline \Bcal_t\} \\
        &\ge \|x_{t, \bar a_t}\|_{V_t^{-1}}\PP_t\{a_t\in \overline \Bcal_t\}
    \end{align*}
    The probability $\PP_t\{a_t\in \overline \Bcal_t\}$ has randomness over algorithm selecting arm from empirical effective Pareto front $\tilde\Ccal_t$, where the set $\overline \Bcal_t$ varies on this random selection. More precisely, the set $\Bcal_t$ and $\overline \Bcal_t$ change as $\wb_t$ changes. But, for any given $\wb_t$, the probability $\PP_t\{a_t\in \overline \Bcal_t\}$ bounds below by the probability that at least one unsaturated arm is evaluated higher compared to all saturated arms, i.e.,  
    \begin{align*}
        \PP_t\{a_t\in \overline \Bcal_t\} 
        &\ge \PP_t\Biggl\{\exists a\in\overline \Bcal_t : x_{t,a}^\top\left(\sum_{\ell\in[L]}w^{(\ell)}_t\tilde\theta\obj{\ell}_{a,t}\right)>\max_{a'\in \Bcal_t}x_{t,a'}^\top\left(\sum_{\ell\in[L]}w^{(\ell)}_t\tilde\theta\obj{\ell}_{a',t}\right)\Biggr\},
    \end{align*}
    where the unsaturated arm exists by $\bar a_*\in \overline \Bcal_t$. Hence this probability bounds below by the probability that the arm $\bar a_*$ is evaluated higher compared to all saturated arms, i.e.,  
    \begin{align*}
        \PP_t\{a_t\in \overline \Bcal_t\} \ge \PP_t\Biggl\{x_{t,\bar a_*}^\top\left(\sum_{\ell\in[L]}w^{(\ell)}_t\tilde\theta\obj{\ell}_{\bar a_*,t}\right)>\max_{a'\in \Bcal_t}x_{t,a'}^\top\left(\sum_{\ell\in[L]}w^{(\ell)}_t\tilde\theta\obj{\ell}_{a',t}\right)\Biggr\}.
    \end{align*}
    On event $\tilde\Ecal_t$, those saturated arms $a'\in\Bcal_t$ satisfy
    \begin{align}\label{eq:2}
         x_{t, \bar a_*}^\top \left(\sum_{\ell\in[L]}w^{(\ell)}_t\theta_*^{(\ell)}\right)-x_{t,a'}^\top\left(\sum_{l\in[L]}w^{(\ell)}_t\theta_*^{(\ell)} \right)  > c_t(\delta)\| x_{t,a'}\|_{V_t^{-1}}
    \end{align}
    and
    \begin{align}\label{eq:3}
        x_{t,a'}^\top \left(\sum_{\ell\in[L]}w^{(\ell)}_t\tilde\theta_{a',t}^{(\ell)}\right)-x_{t,a'}^\top\left(\sum_{l\in[L]}w^{(\ell)}_t\theta_*^{(\ell)} \right) \le c_t(\delta)\| x_{t,a'}\|_{V_t^{-1}}.
    \end{align}
    Subtracting \cref{eq:3} from \cref{eq:2}, we get
    \begin{align*}
        x_{t, \bar a_*}^\top \left(\sum_{\ell\in[L]}w^{(\ell)}_t\theta_*^{(\ell)}\right)-x_{t,a'}^\top \left(\sum_{\ell\in[L]}w^{(\ell)}_t\tilde\theta_{a',t}^{(\ell)}\right) \ge 0.
    \end{align*}
    for all $a'\in\Bcal_t$. Using this inequality, the Probability bounds as
    \begin{align*}
        \PP_t\{a_t\in \overline \Bcal_t\} &\ge \PP_t\Biggl\{x_{t,\bar a_*}^\top
            \left(\sum_{\ell\in[L]}w\obj{\ell}_t\tilde\theta\obj{\ell}_{\bar a_*,t}\right)
            >x_{t,\bar a_*}^\top
            \left(\sum_{\ell\in[L]}w^{(\ell)}_t\theta_*^{(\ell)}\right),
            \tilde\Ecal_t\Biggr\}\\
            &\ge \PP_t\Biggl\{x_{t,\bar a_*}^\top \left(\sum_{\ell\in[L]}w\obj{\ell}_t\tilde\theta\obj{\ell}_{\bar a_*, t}\right)
        >x_{t,\bar a_*}^\top\left(\sum_{\ell\in[L]}w^{(\ell)}_t\theta_*^{(\ell)}\right)\Biggr\}-(1-\mathbb P_t\{\tilde\Ecal_t\}) 
    \end{align*}
    Since $\tilde\theta\obj{\ell}_{\bar a_*, t}$ is objective wise independent, the probability bounds by objective wise probability,
    \begin{align*}
        &\PP_t\Biggl\{x_{t,\bar a_*}^\top \left(\sum_{\ell\in[L]}w\obj{\ell}_t\tilde\theta\obj{\ell}_{\bar a_*, t}\right)
        >x_{t,\bar a_*}^\top\left(\sum_{\ell\in[L]}w^{(\ell)}_t\theta_*^{(\ell)}\right)\Biggr\}
        \ge \bigcap_{\ell\in[L]} 
            \PP_t\{x_{t,\bar a_*}^\top\tilde\theta\obj{\ell}_{\bar a_*, t}
                >x_{t,\bar a_*}^\top\theta_*^{(\ell)}\}
    \end{align*}
    Removing the random vector $\wb_t$, this inequality holds for any $\wb_t$. In other words, this inequality holds for any random selection of arms from the set $\tilde\Ccal_t$. Hence, we get
    \begin{align*}
        \PP_t\{a_t\in \overline \Bcal_t\} &\ge \bigcap_{\ell\in[L]} 
            \PP_t\{x_{t,\bar a_*}^\top\tilde\theta\obj{\ell}_{\bar a_*, t}
                >x_{t,\bar a_*}^\top\theta_*^{(\ell)}\}
            -(1-\mathbb P_t\{\tilde\Ecal_t\}) \\
        &\ge \bigcap_{\ell\in[L]} 
            \PP_t\{x_{t,\bar a_*}^\top\tilde\theta\obj{\ell}_{\bar a_*, t}
                >x_{t,\bar a_*}^\top\theta_*^{(\ell)}\}
            -\frac{\delta}{T} \\
        &= \PP_t\{x_{t,\bar a_*}^\top\tilde\theta\obj{1}_{\bar a_*, t}
                >x_{t,\bar a_*}^\top\theta_*^{(1)}\}^L
            -\frac{\delta}{T}
    \end{align*}
    As we remove $\wb_t$, the probability $\PP_t\{a_t\in \overline \Bcal_t\}$ gets exponentially small as the number of objective increases. We remove this by adopting optimistic sampling strategy. With the number multiple samples $M$, following \cref{lem:opt}, we have
    \begin{align*}
        \PP_t\{x_{t,\bar a_*}^\top\tilde\theta\obj{1}_{\bar a_*, t}
                >x_{t,\bar a_*}^\top\theta_*^{(1)}\} \ge 1-(1-p)^M.
    \end{align*}
    Hence, we get
    \begin{align*}
        \PP_t\{a_t\in \overline \Bcal_t\} \ge (1-(1-p)^M)^L-\frac{\delta}{T}
    \end{align*}
    This inequality holds for any $\wb_t$. With $M =\lceil 1-\frac{\log L}{\log (1-p)}\rceil$, we have $(1-(1-p)^M)^L \ge p$. Finally, we have 
    \begin{align*}
        \EE_t[\|x_{a_t}\|_{V_t^{-1}}] &\ge \|x_{ \bar a_t}\|_{V_t^{-1}}\left(p-\frac{\delta}{T}\right)
    \end{align*}
    Replacing the term $\|x_{ \bar a_t}\|_{V_t^{-1}}$ to $\|x_{a_t}\|_{V_t^{-1}}$, we get.
    \begin{align*}
        \EE_t[\Delta_{a_t}^{EPR}]\le \left(1+\frac{2}{p-\frac{\delta}{T}}\right)c_t(\delta)\EE_t[\|x_{ a_t}\|_{V_t^{-1}}] + \frac{\delta}{T}\Delta_{max}
    \end{align*}
    
\end{proof}

Now we begin the proof of \cref{thm:regret}.

\begin{proof}    
    We have
    \begin{align*}
        \EE[EPR(T)] 
        &= \sum_{t=1}^T\EE[\Delta_{a_t}^{EPR}] \\
        &= \PP(\hat\Ecal)\sum_{t=1}^T\EE[\Delta_{a_t}^{EPR}\ind\{\hat\Ecal\}] + (1-\PP(\hat\Ecal))\Delta_{\text{max}} \\
        &\le \sum_{t=1}^T\EE[\Delta_{a_t}^{EPR}\ind\{\hat\Ecal\}] + \delta \Delta_{\text{max}} \\
        &= \sum_{t=1}^T\EE[\EE_t[\Delta_{a_t}^{EPR}]\ind\{\hat\Ecal\}] + \delta \Delta_{\text{max}} 
    \end{align*}

    By \cref{lem:bound}, bounding the term $\EE_t[\Delta_{a_t}^{EPR}]$, we have
    \begin{align*}
        \EE[EPR(T)] &\le \sum_{t=1}^T\left(1+\frac{2}{0.15-\frac{\delta}{T}}\right)c_t(\delta)\EE[\EE_t[\|x_{t, a_t}\|_{V_t^{-1}}]\ind\{\hat\Ecal\}] + 2\delta \Delta_{\text{max}} \\
        &\le \left(1+\frac{2}{0.15-\frac{\delta}{T}}\right)c_T(\delta)\EE[\sum_{t=1}^T\EE_t[\|x_{t, a_t}\|_{V_t^{-1}}]\ind\{\hat\Ecal\}] + 2\delta \Delta_{\text{max}} \\
        &= \left(1+\frac{2}{0.15-\frac{\delta}{T}}\right)c_T(\delta)\EE[\sum_{t=1}^T\|x_{t, a_t}\|_{V_t^{-1}}] + 2\delta \Delta_{\text{max}} \\
        &\le \left(1+\frac{2}{0.15-\frac{\delta}{T}}\right)c_T(\delta)\sqrt{2Td\log\left(1+\frac{T}{\lambda}\right)} + 2\delta \Delta_{\text{max}},
    \end{align*}
    where the last inequality follows by \cref{prop:abbasi1},
\end{proof}
\section{Additional Technical Tools}

\begin{proposition}[Gordan's Theorem, page 31, \citealt{mangasarian1994nonlinear}]\label{lem:Gordan}
    For given matrix $M\in\RR^{m\times L}$, either one of the following statements holds, but not both.
    \begin{enumerate}
        \item There exists $L$-dimensional vector $\wb$, that $M\wb$ has all positive entries.
        \item $M^\top \beta = 0$, $\beta\succ 0$ has solution $\beta \in \RR^m$.
    \end{enumerate}
\end{proposition}

\begin{proposition}[Lemma 11, \citealt{abbasi2011improved}]\label{prop:abbasi1}
    Let $\lambda\ge1$. For arbitrary sequence $(x_{t, a_t})_{t\in[T]}$, we have
    \begin{equation*}
        \sum_{t=1}^T\|x_{t, a_t}\|_{V_t^{-1}}^2 \le 2d\log\left(1+\frac{T}{\lambda}\right).
    \end{equation*}
\end{proposition}

\begin{lemma}[Theorem 2, \citealt{abbasi2011improved}]\label{lem:abbasi1}
Let $(\Fcal_t)_{t\ge0}$ be a filtration. Let $(\xi_t\obj{\ell})$ be a real-valued stochastic process such that $\xi_t\obj{\ell}$ is conditionally $R$-sub-Gaussian, given filtration $\Fcal_t$ for any $\ell \in[L]$. Then with probability at least $1-\delta$, the event
\begin{equation*}
    \hat\Ecal_t =\Biggl\{\forall \ell\in[L]:\|\hat\theta_t\obj{\ell}-\theta_*\obj{\ell}\|_{V_t}\le R\sqrt{d\log\left(\frac{1+(t-1)/(\lambda d)}{\delta/L}\right)}+\lambda^{1/2}\Biggr\} 
\end{equation*}
holds for all $t\ge1$.
\end{lemma}
\begin{proof}
    By Theorem 2 in \citet{abbasi2011improved}, and union bound with $L$.
\end{proof}

\begin{lemma}[Definition 1, \citealt{abeille2017linear}] \label{lem:abeille1}
On event $\hat\Ecal_t$, with probability at least $1-\delta$, all sampled parameters $(\tilde\theta_{t,m}\obj{\ell})_{m\in[M], \ell\in[L]}$ follow concentration property, i.e., 

    \begin{align*}
        \tilde\Ecal_t &:= \Biggl\{\|\tilde\theta_{t,m}\obj{\ell}-\hat\theta_t\obj{\ell}\|_{V_t}\le \sqrt{2d\log\frac{2LMd}{\delta}}\left(R\sqrt{d\log\left(\frac{1+(t-1)/(\lambda d)}{\delta/L}\right)}+\lambda^{1/2}\right)\Biggr\}.
    \end{align*} for all $m\in[M], \ell \in[L]$.
\end{lemma} 
\begin{proof}
    By Definition 1 in \citet{abeille2017linear}, and union bound with $M$ and $L$.
\end{proof}

\section{Discussions}\label{app:limitation}
Our proposed algorithm demonstrates strong empirical performance with various settings. But its theoretical worst-case regret bound is not tighter than that of UCB-based algorithms. This gap between UCB-type algorithms and TS algorithms is well-known in the regret analysis of previous TS algorithms~\cite{agrawal2013thompson, abeille2017linear} bounding the worst-case frequentist regret. 

Our work is studied under the standard linear contextual bandit setting. Our framework can be readily extended to generalized linear contextual bandits.
Extension to more complex function class such as neural networks requires analysis that is beyond the scope of this work, but is certainly the promising avenue for future work.

Yet, our work introduces the first randomized algorithm with Pareto regret guarantees in the multi-objective bandit framework. We hope that our work lays a foundation basis for extending such techniques to follow-up works.

\section{Computing resources for experiments}\label{app:resources}
All experiments are conducted with INTEL(R) XEON(R) GOLD 6526Y CPU and 4 TB memory. The software environment includes Python 3.12.7, Scipy 1.14.1, and Numpy 1.26.4. The experiments took approximately 4 hours to 1 day, as it takes longer with increasing numbers of arms, dimensions, and objectives.

\newpage

\section{Additional experimental results}\label{app:result}

\begin{figure}[htbp!]
    \centering
    \includegraphics[width=0.90\linewidth]{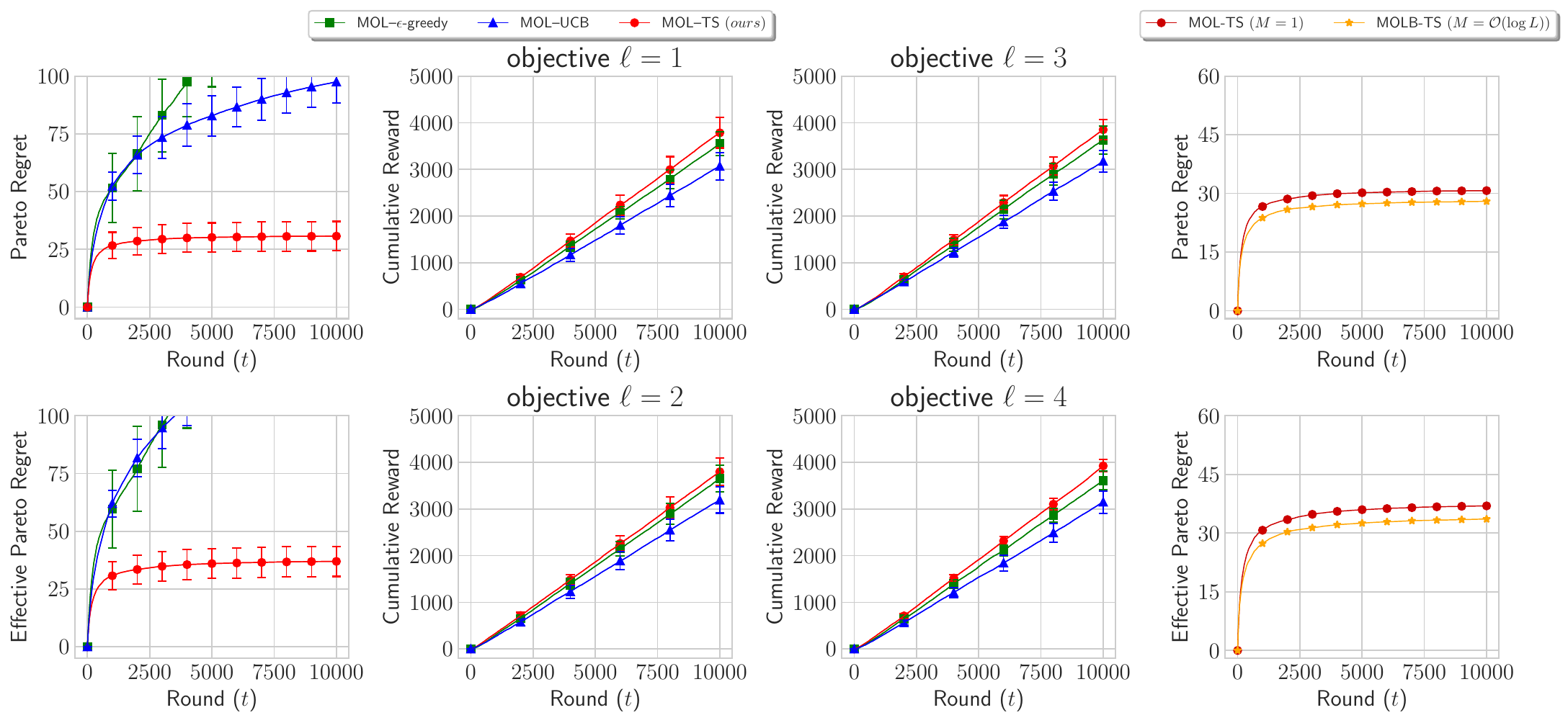}
    \caption{Experimental results with $K = 50,~ d=10,~L=4$}
\end{figure}
\begin{figure}[htbp!]
    \centering
    \includegraphics[width=0.90\linewidth]{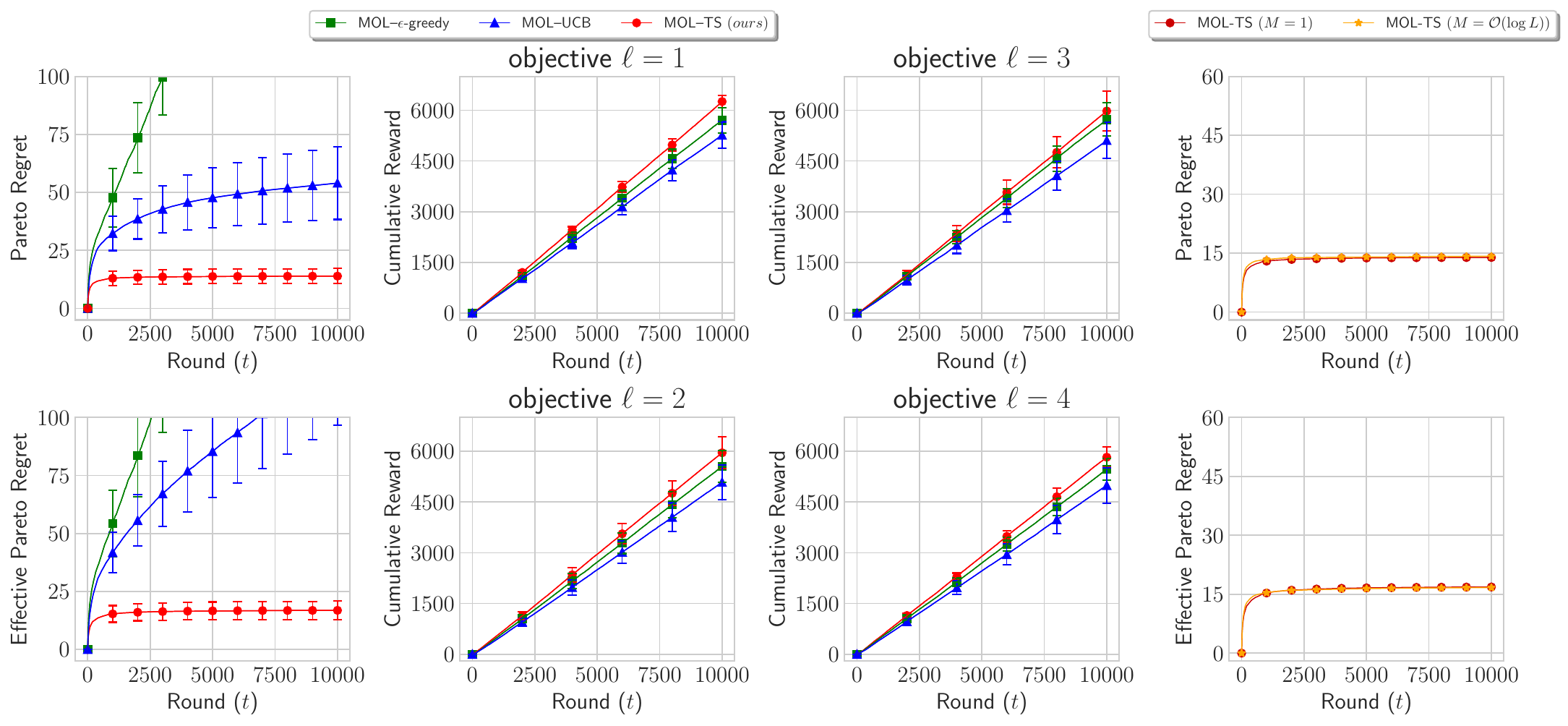}
    \caption{Experimental results with $K = 100,~ d=5,~L=4$}
\end{figure}
\begin{figure}[htbp!]
    \centering
    \includegraphics[width=0.9\linewidth]{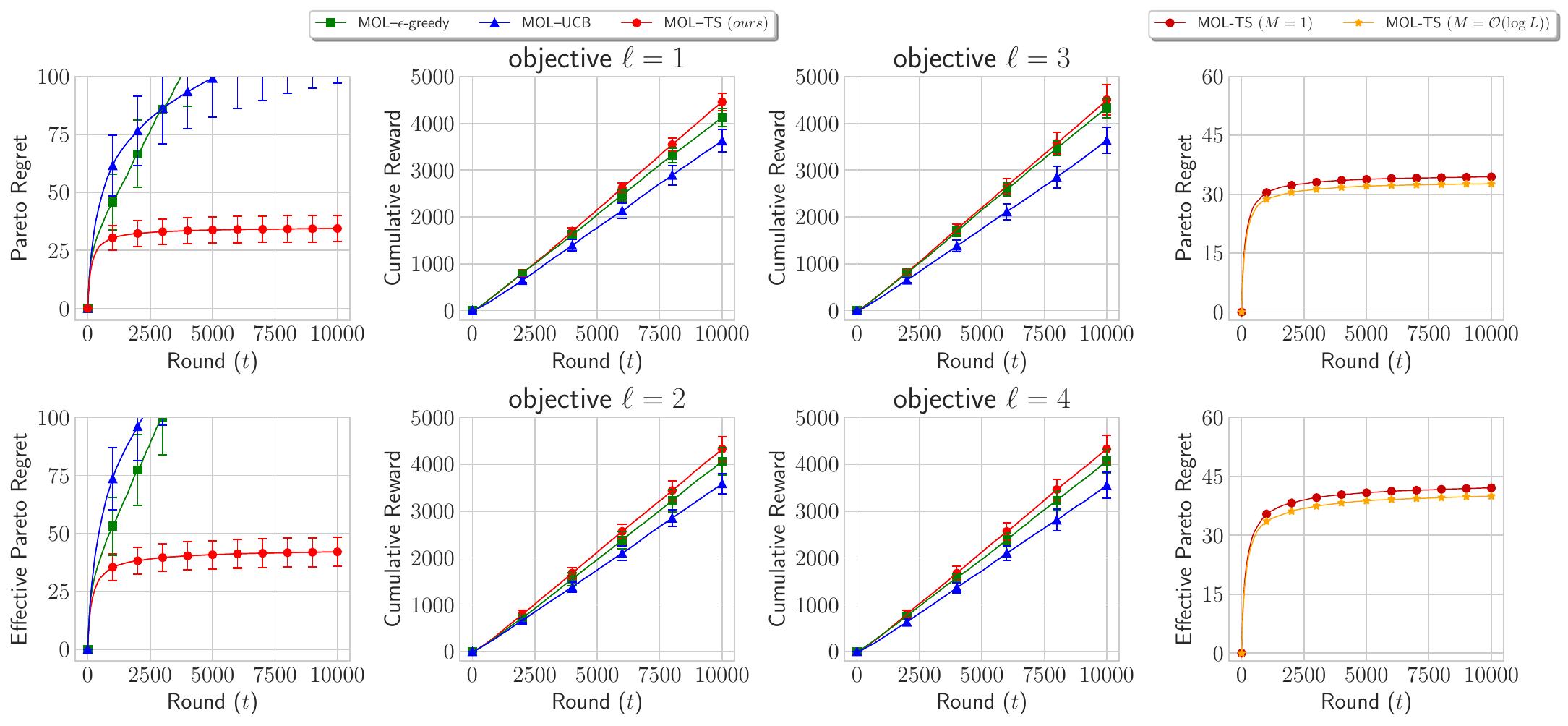}
    \caption{Experimental results with $K = 100,~ d=10,~L=4$}
\end{figure}
\begin{figure}
    \centering
    \includegraphics[width=0.9\linewidth]{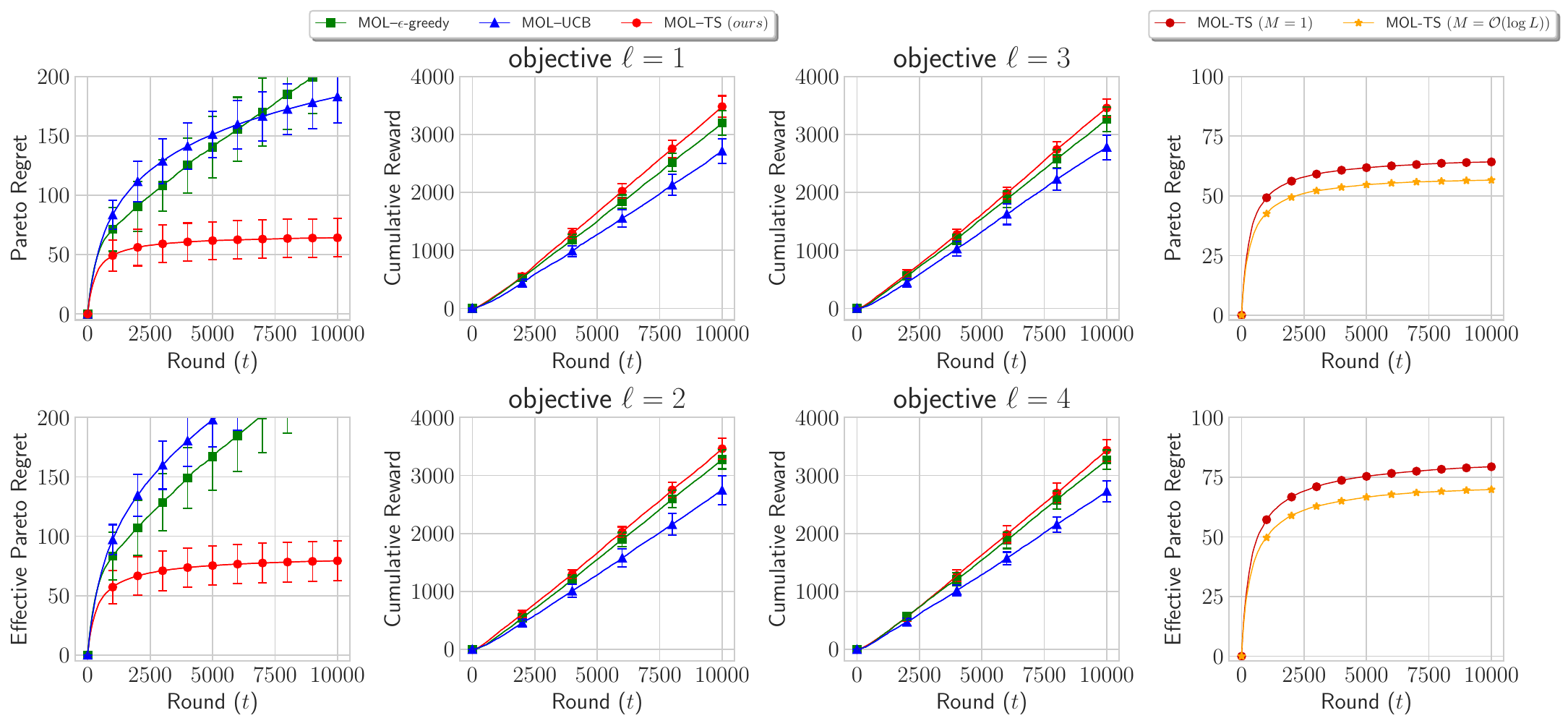}
    \caption{Experimental results with $K = 100,~ d=15,~L=4$}
\end{figure}
\begin{figure}
    \centering
    \includegraphics[width=0.9\linewidth]{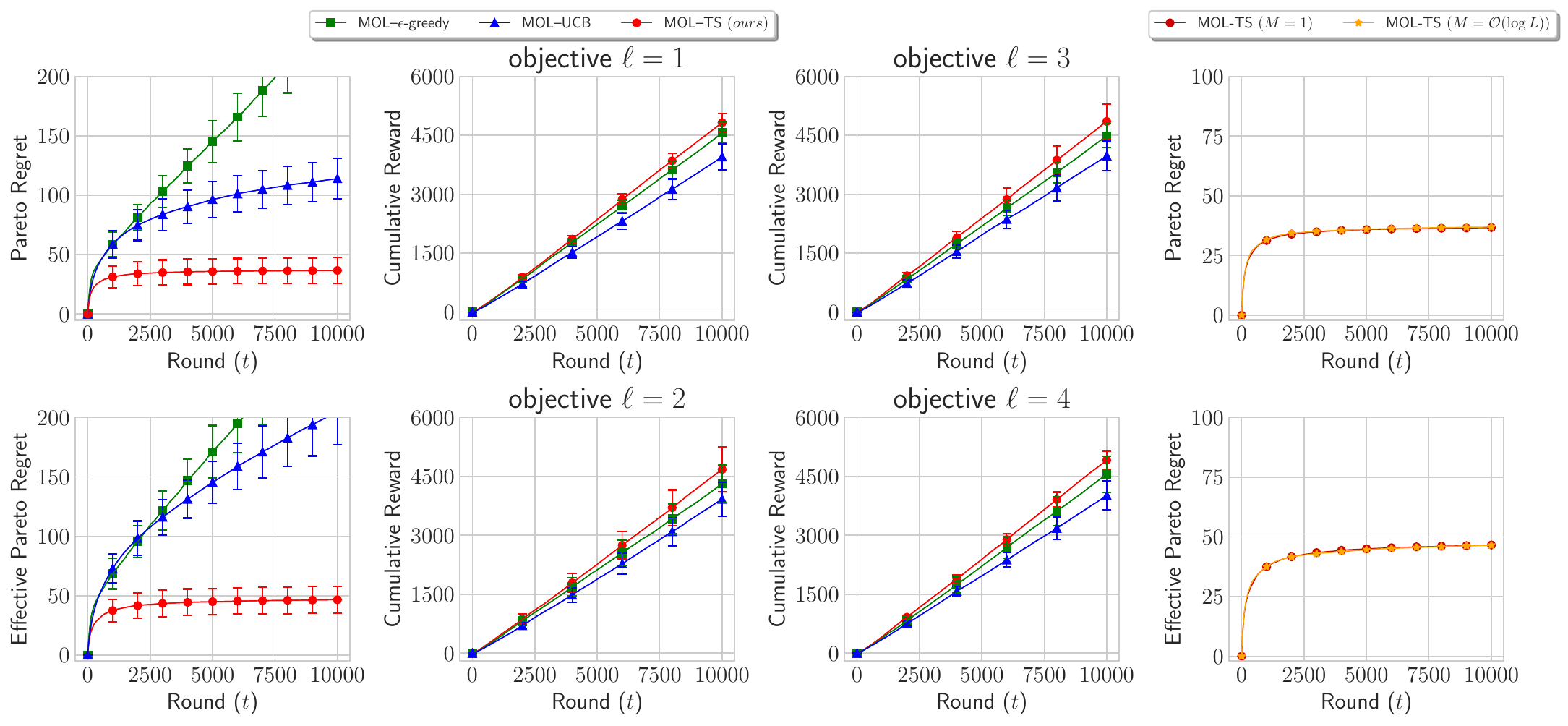}
    \caption{Experimental results with $K = 200,~ d=10,~L=4$}
\end{figure}
\begin{figure}
    \centering
    \includegraphics[width=0.9\linewidth]{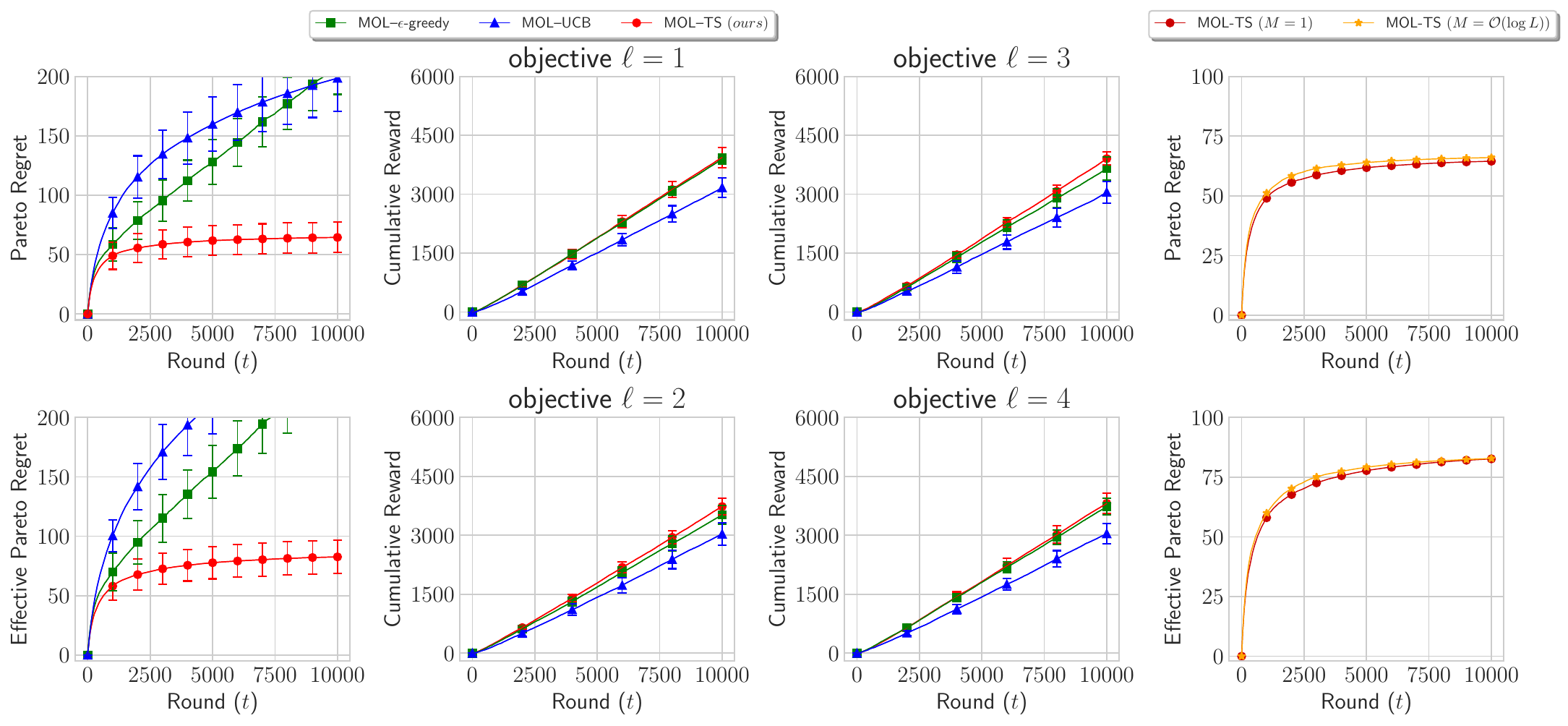}
    \caption{Experimental results with $K = 200,~ d=15,~L=4$}
\end{figure}

\begin{figure}
    \centering
    \includegraphics[width=0.95\linewidth]{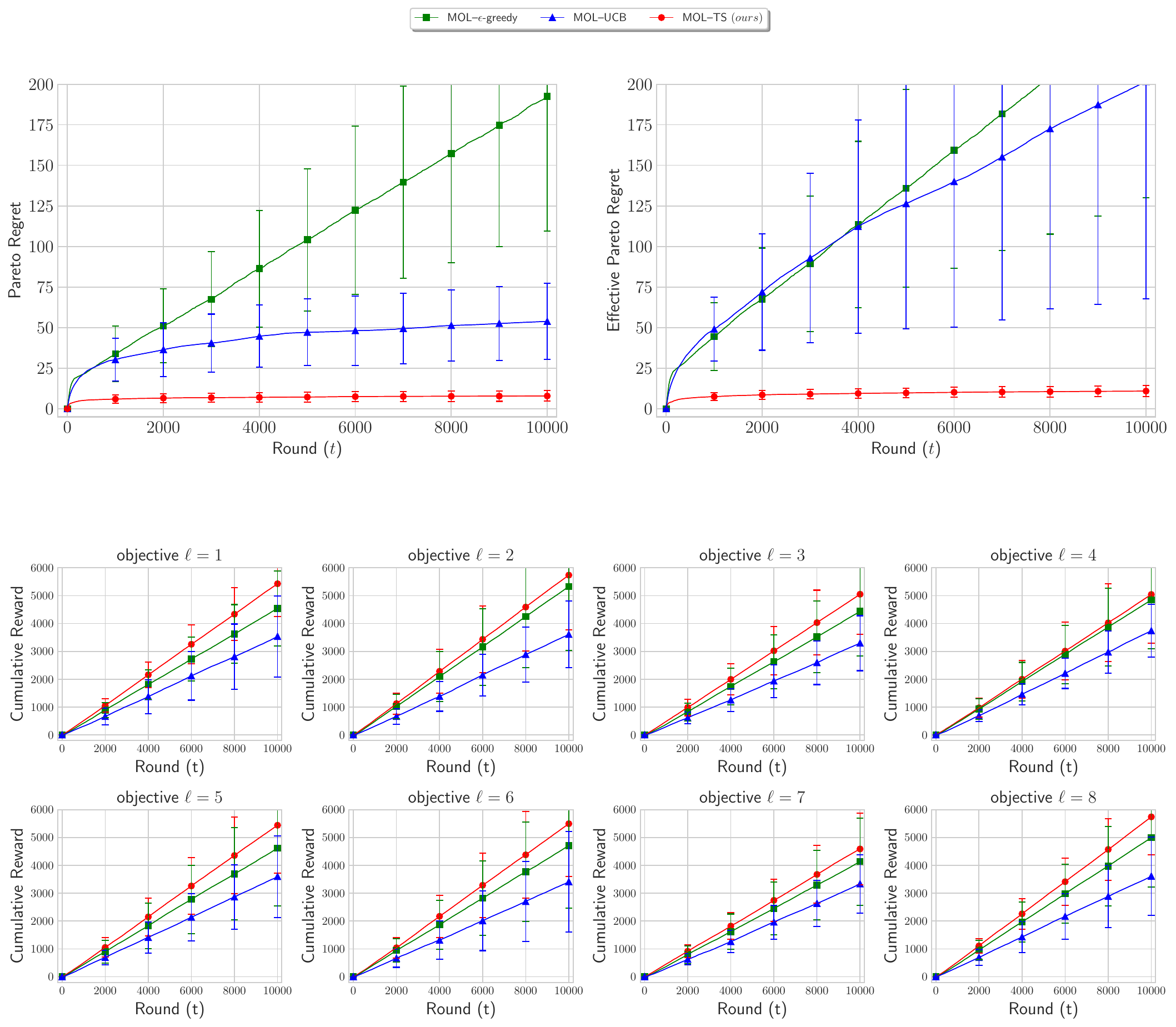}
    \caption{Experimental results with $K = 50,~ d=5,~L=8$}
\end{figure}
\begin{figure}
    \centering
    \includegraphics[width=0.95\linewidth]{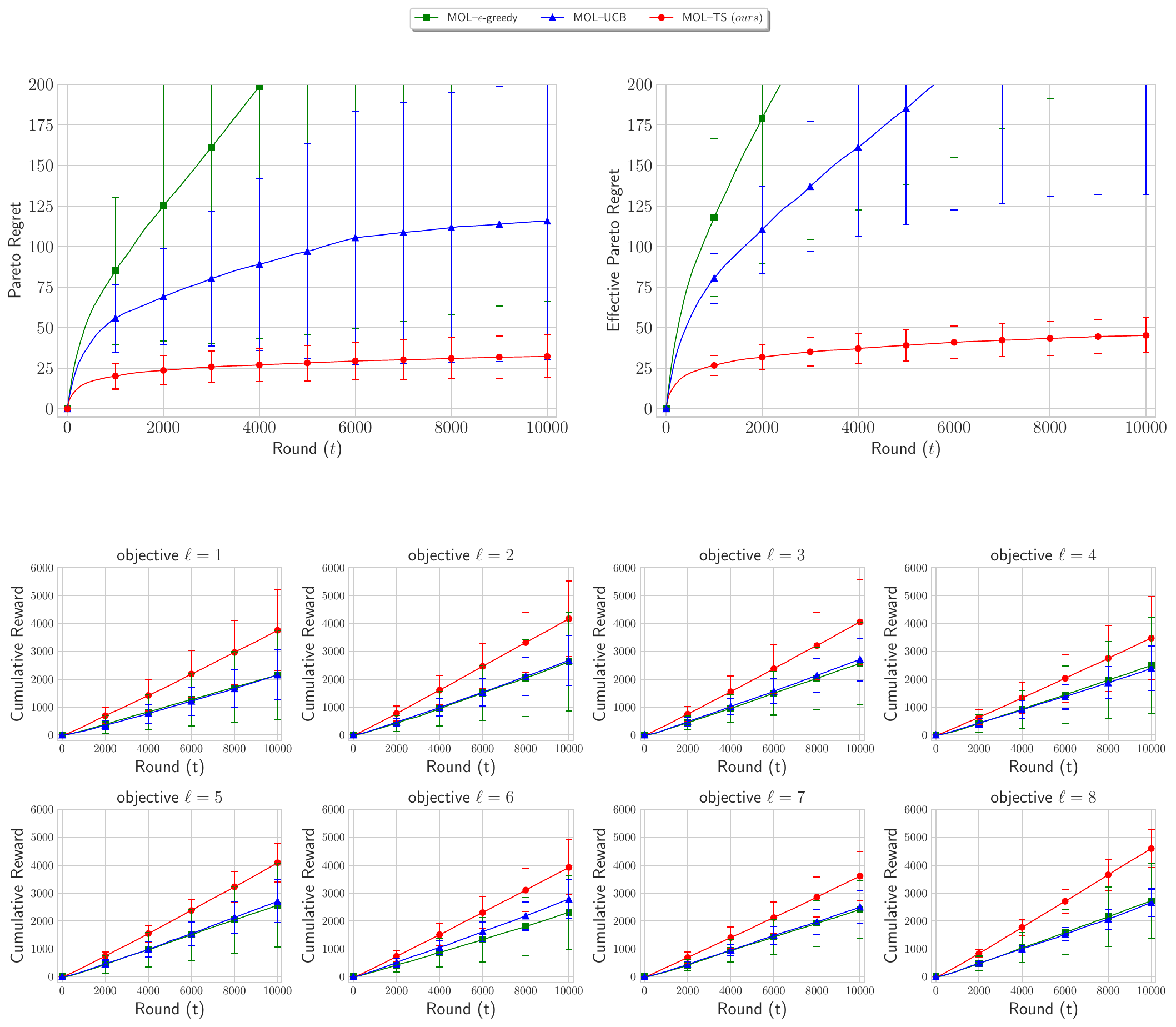}
    \caption{Experimental results with $K = 100,~ d=10,~L=8$}
\end{figure}
\begin{figure}
    \centering
    \includegraphics[width=0.95\linewidth]{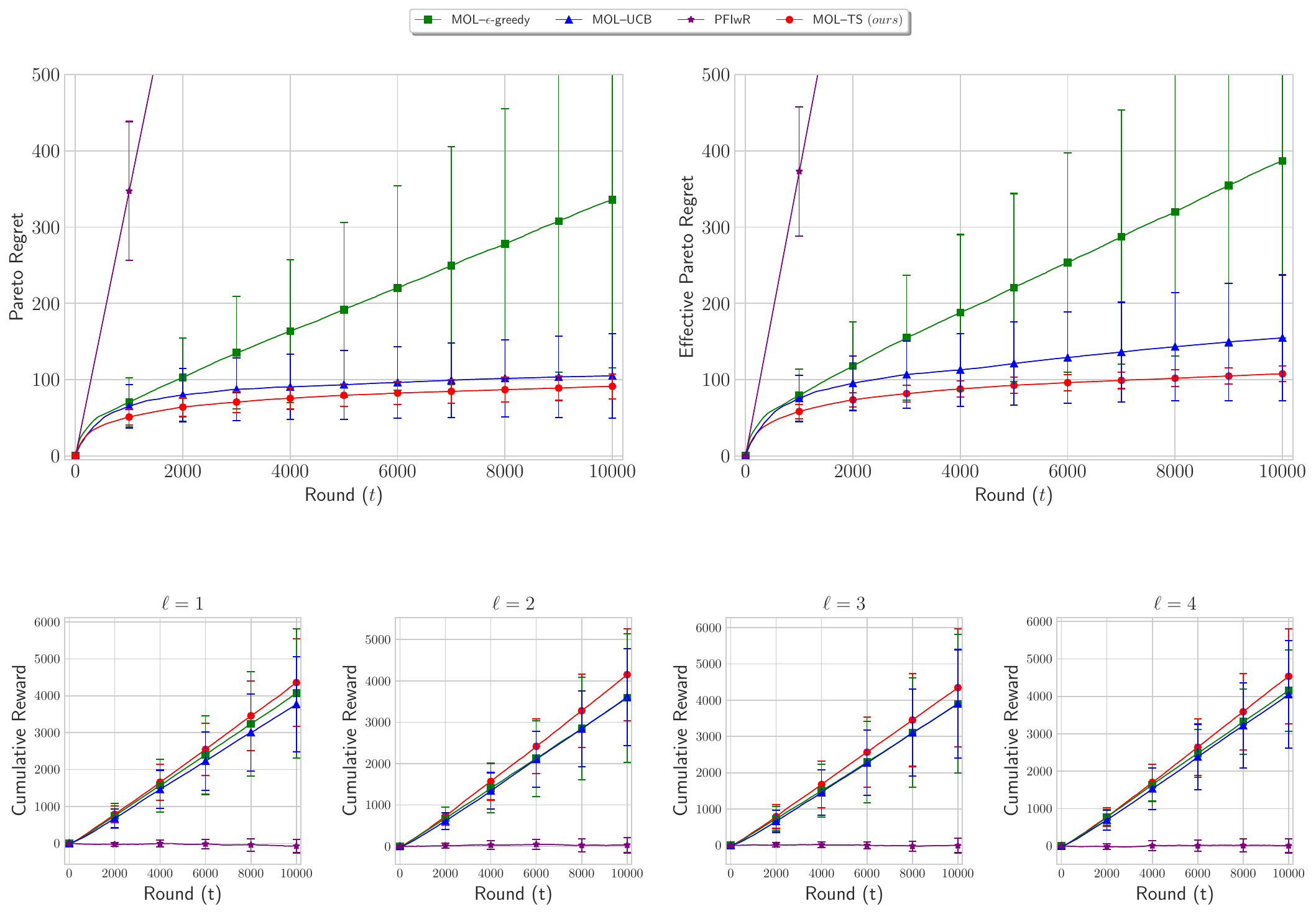}
    \caption{Experimental results with $K = 100,~ d=10,~L=4$, linear (non-contextual) setting.}
\end{figure}
\begin{figure}
    \centering
    \includegraphics[width=0.95\linewidth]{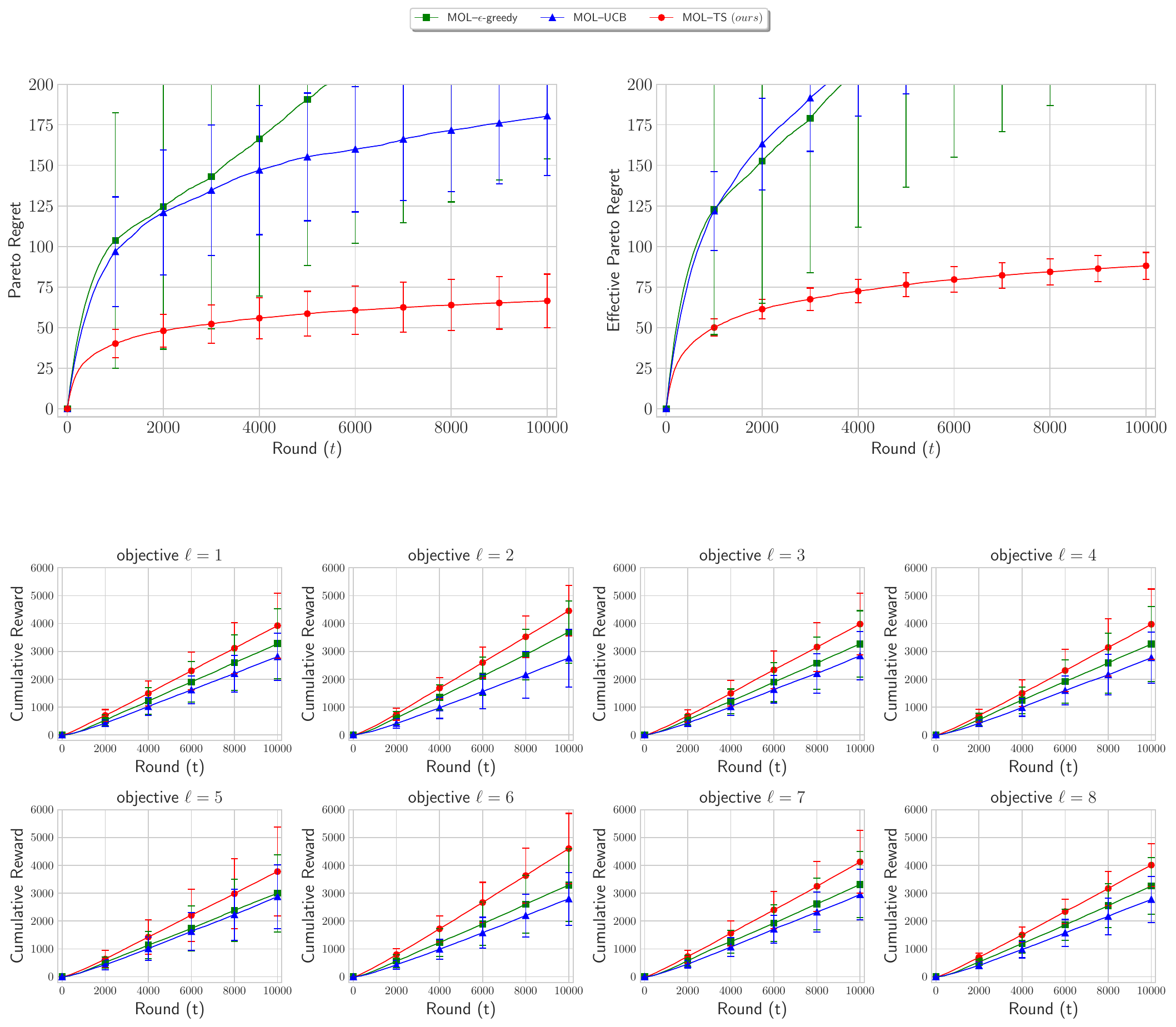}
    \caption{Experimental results with $K = 200,~ d=15,~L=8$}
\end{figure}
\begin{figure}
    \centering
    \includegraphics[width=0.95\linewidth]{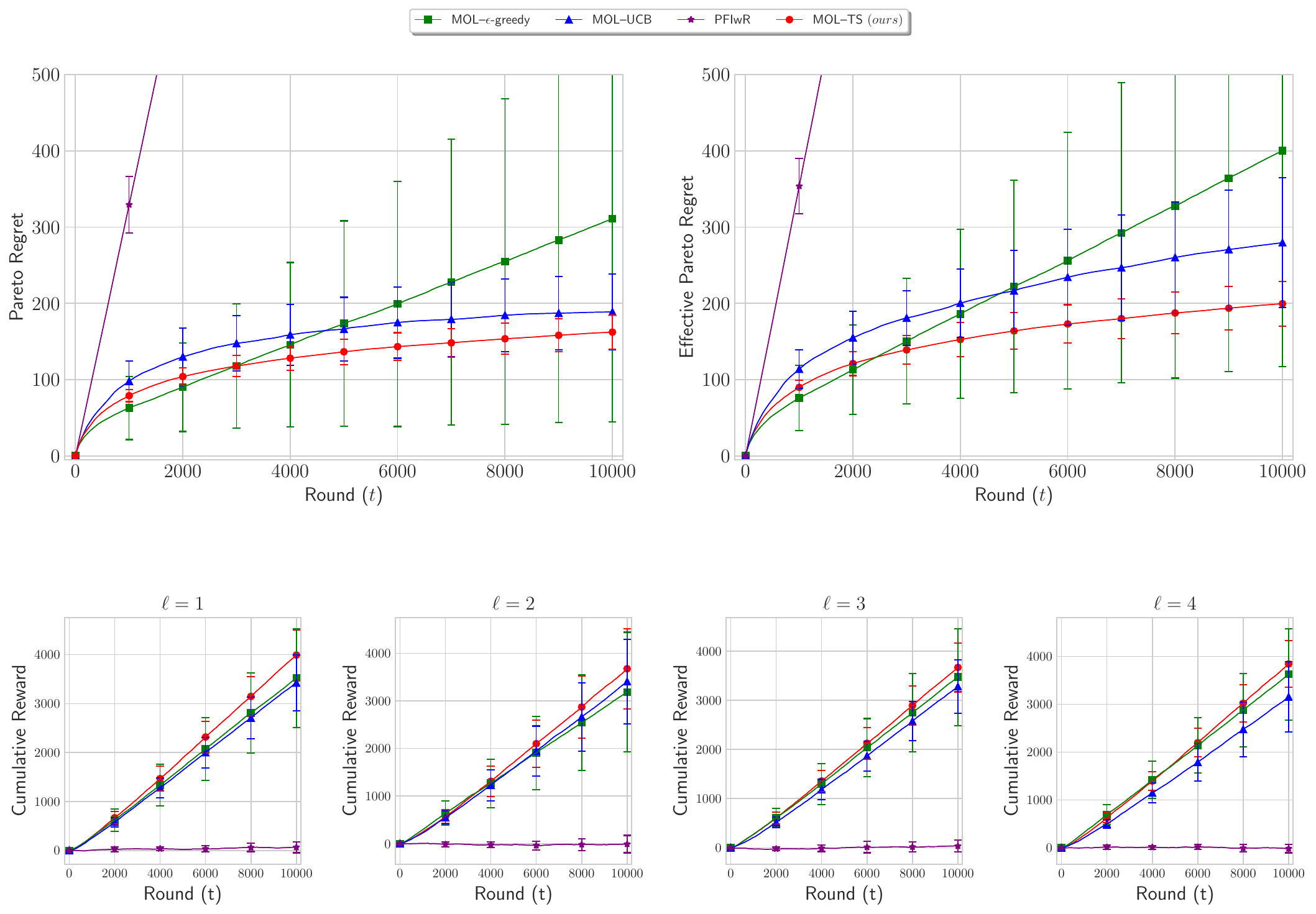}
    \caption{Experimental results with $K = 200,~ d=15,~L=4$, linear (non-contextual) setting.}
\end{figure}

\end{document}